\documentclass[journal]{IEEEtran}
\ifCLASSINFOpdf                 %
  \usepackage[pdftex]{graphicx}
  \DeclareGraphicsExtensions{.pdf,.jpeg,.png}
\graphicspath{{figures/}}
\usepackage{amsmath}
\usepackage{amssymb}
\usepackage{bm}

\ifCLASSOPTIONcompsoc
 \usepackage[caption=false,font=normalsize,labelfont=sf,textfont=sf]{subfig}
\else
 \usepackage[caption=false,font=footnotesize]{subfig}
\fi
\newcommand{\real}{{\mathbb{R}}}

\usepackage{tikz}
\usetikzlibrary{shapes,arrows,positioning}

\tikzset{
    >=stealth',
    punkt/.style={
           rectangle,
           rounded corners,
           draw=black, very thick,
           text width=7.5em,
           minimum height=2em,
           text centered},
    pil/.style={
           ->,
           thick,
           shorten <=2pt,
           shorten >=2pt,}
}
\tikzstyle{block} = [rectangle, draw, thick,fill=blue!20, 
    text width=6em, rounded corners, text centered, minimum height=4em]
\tikzstyle{block2} = [rectangle, draw, thick, 
    text width=6em, rounded corners, text centered, minimum height=4em]
\tikzstyle{line} = [draw, -latex',text width=4.5em,]
\tikzstyle{cloud} = [draw, ellipse,fill=red!20, node distance=3cm,
    minimum height=2em]

\begin{document}
%
\title{Ergodic Exploration of Distributed Information}
%
%
%

\author{Lauren~M.~Miller,$^{\ast1}$
        Yonatan~Silverman,$^{1}$
        Malcolm~A.~MacIver,$^{1,2}$
        and~Todd~D.~Murphey$^{1}$
\thanks{$^{1}$Department of Mechanical Engineering, $^{2}$Department
    of Biomedical Engineering, Northwestern University, Evanston,
                Illinois, USA (
$^\ast$e-mail:
     LMiller@u.northwestern.edu)}
\thanks{This work was supported by Army Research
Office grant W911NF-14-1-0461, NSF grants IOB-0846032, CMMI-0941674,
CMMI-1334609 and Office of Naval Research Small Business Technology
Transfer grants N00014-09-M-0306 and N00014-10-C0420 to M.A.M.}
}

\maketitle

\begin{abstract}
  This paper presents an active search trajectory synthesis technique
  for autonomous mobile robots with nonlinear measurements and dynamics.
  The presented approach uses the ergodicity of a planned trajectory
  with respect to an expected information density map to close the loop
  during search. The ergodic control algorithm does not rely on
  discretization of the search or action spaces, and is well posed for
  coverage with respect to the expected information density whether the
  information is diffuse or localized, thus trading off between
  exploration and exploitation in a single objective function. As a
  demonstration, we use a robotic electrolocation platform to estimate
  location and size parameters describing static targets in an
  underwater environment. Our results demonstrate that the ergodic
  exploration of distributed information (EEDI) algorithm outperforms
  commonly used information-oriented controllers, particularly when
  distractions are present.
\end{abstract}


%
\IEEEpeerreviewmaketitle

%
%
%
%

\section{Introduction}
\label{intro}
\IEEEPARstart{I}{n} the context of exploration, ergodic trajectory
optimization computes control laws that drive a dynamic system along
trajectories such that the amount of time spent in regions of the state
space is proportional to the expected information gain in those regions.
Using ergodicity as a metric encodes both exploration and
exploitation---both the need for nonmyopic search when variance is high
and convexity is lost, as well as myopic search when variance is low and
the problem is convex. By encoding these needs into a metric
\cite{Mathew}, generalization to nonlinear dynamics is possible using
tools from optimal control. We show here that different dynamical
systems can achieve nearly identical estimation performance using EEDI.
                    
\begin{figure}[!b] \centering \vspace{-10pt} 
  \subfloat[The SensorPod robot (white cylinder at end of vertical Z stage) is used
  to demonstrate  EEDI for active search.] {\label{FishB}\includegraphics[trim=.18in .4in
    .02in .33in,clip=true,width= \columnwidth  ]{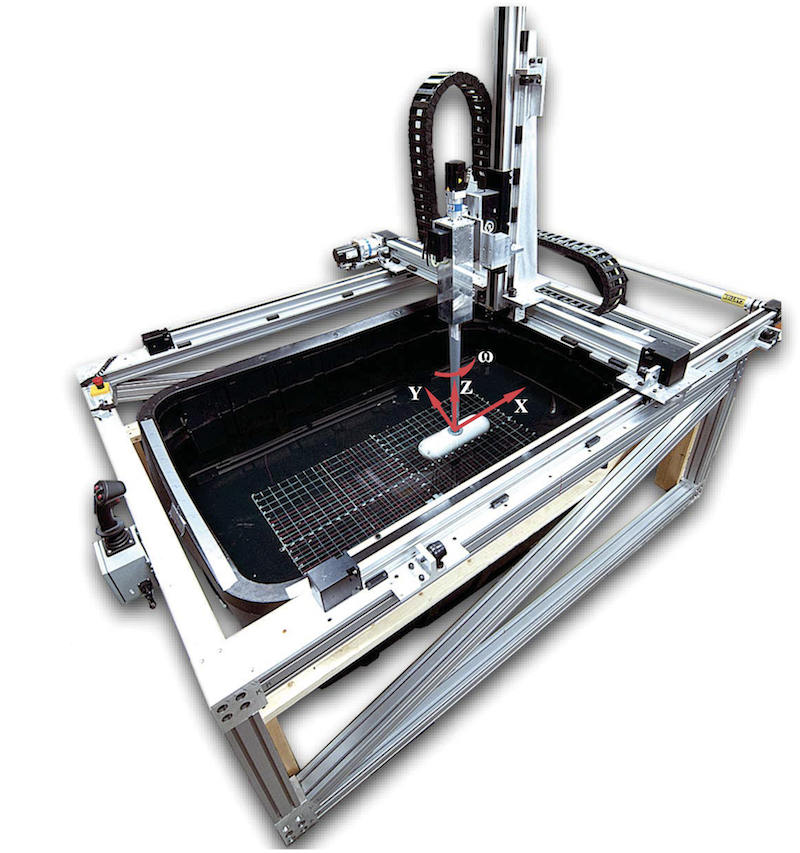}}\\
\subfloat[\emph{Apteronotus albifrons} (photograph courtesy
  of Per Erik Sviland.) ]
  {\label{FishA}\includegraphics[width=  \columnwidth ]{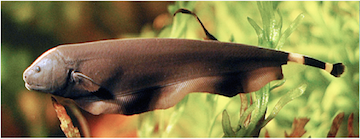}}
  \caption{ The SensorPod  (a) uses a sensing modality 
   inspired by weakly electric fish such as the black ghost
  knifefish (b). The SensorPod  is mounted on a 4DOF gantry and submerged within a 1.8 m x 2.4 m
  x 0.9 m (l,w,h) tank (see multimedia attachment).
  }
\label{FIG}
\end{figure}

The SensorPod robot (Fig. \ref{FishB}), which we use as a motivating
example and an experimental platform in Section \ref{results}, measures
disturbances in a self-generated, weak electric field. This sensing
modality, referred to as electrolocation, is inspired by a type of
freshwater tropical fish (Fig. \ref{FishA}, \cite{Krah13a,
  Nels06a,Neve13a}), and relies on the coupled emission and detection of
a weak electric field. Electrolocation is ideally suited for low
velocity, mobile vehicles operating in dark or cluttered environments
\cite{Solb08a,MacI04a,Neve13a}. The sensing range for electrolocation is
small, however, so the fish or robot must be relatively close to an
object to localize it. Also, as the sensors are rigid with respect to
the body, the movement of those sensors involves the dynamics of the
entire robot. As we will see in Section \ref{expsetup}, the measurement
model for electrolocation is also highly nonlinear and the dynamics of
both biological fish and underwater robots are generally nonlinear.
Consideration of sensor physics and robot dynamics when planning
exploration strategies is therefore particularly important. The same
applies to many near-field sensors such as tactile sensors,
ultra-shortwave sonar, and most underwater image sensors (e.g.
\cite{Cowen97}). Experiments carried out using the SensorPod robot
demonstrate that the ergodic exploration of distributed information
(EEDI) algorithm is successful in several challenging search scenarios
where other algorithms fail.

\medbreak
  \noindent The contributions of this paper can be summarized as
  follows:
\begin{enumerate}
\item application of ergodic exploration for general, nonlinear,
  deterministic control systems to provide closed-loop coverage with
  respect to the evolving expected information density, and
\item   validation of ergodic search in an experimental and simulated 
  underwater sensing setting. We demonstrate both that ergodic search performs as well as alternative algorithms in nominal scenarios, and that ergodic search outperforms alternatives  when distractors are present.
\end{enumerate} 

Section \ref{relatedwork} begins with a discussion of related work.
Ergodicity as an objective for active sensing is presented in Section
\ref{ergodicitydiscussion}, including the benefits and distinguishing
features of ergodic trajectory optimization. Section \ref{trajopt}
includes an overview of ergodic trajectory optimization. In Section
\ref{expsetup}, we describe the SensorPod experimental platform and
nonlinear measurement model, and introduce the stationary target
localization task used to demonstrate EEDI. We also discuss the
components of closed-loop EEDI for target localization using the
SensorPod in Section \ref{expsetup}. In Section \ref{results}, we
present data from multiple estimation scenarios, including comparison to
several alternative algorithms, and closed-loop EEDI implementation
using different dynamic models for the SensorPod. We also include a
multimedia video attachment with an extended description of the
SensorPod platform and measurement model used in Sections \ref{expsetup}
and \ref{results}, and an animated overview of the steps of the EEDI
algorithm for this system.

\section{Motivation \& related work}
\label{relatedwork}
The ability to actively explore and respond to uncertain scenarios is
critical in enabling robots to function autonomously. In this paper, we
examine the problem of control design for mobile sensors carrying out
active sensing tasks. Active sensing \cite{kreucher05s, Fox98} or sensor
path planning \cite{Cai2009}, refers to control of sensor parameters,
such as position, to acquire information or reduce uncertainty.
Applications include prioritized decision making during search and
rescue \cite{Toh06, Cooper08}, inspection for flaws
\cite{hollinger2013}, mine detection \cite{Cai2009}, object
recognition/classification \cite{Denzler02, Arbel99, Ye99},
next-best-view problems for vision systems \cite{vazquez2001,
  massios1998, takeuchi1998}, and environmental modeling/field
estimation \cite{Cao2013, bender2013,marchant2014}. Planning for
search/exploration is challenging as the planning step necessarily
depends not only on the sensor being used but on the quantity being
estimated, such as target location versus target size. Methods for
representing and updating the estimate and associated uncertainty---the
belief state---and a way of using the belief state to determine expected
information are therefore required.

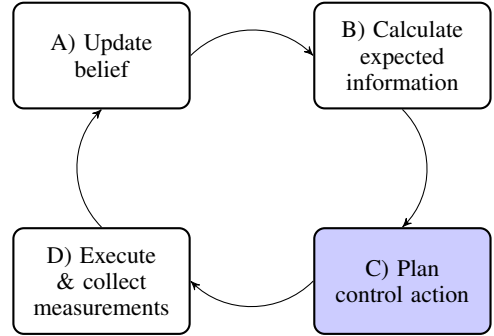
\begin{figure}[t]
\centering
\begin{tikzpicture}
  \tikzstyle{every node}=[font=\small] [node distance = 2cm,
  auto]

  \node [block2] (EID) {A) Update belief };

  \node [block2, right of=EID,node distance=4cm] (control) {B) Calculate
    expected information };

  \node [block,below of=control,node distance=3cm] (explore) {C) Plan
    control action };

  \node [block2, left of=explore, node distance=4cm] (update) {D)
    Execute \& collect measurements};

  \path[->] (EID.east) edge[bend left=45] node[above] {} (control.west);

  \path[->] (control.south) edge[bend
  left=45]
  (explore.north);

  \path[->] (explore.west) edge[bend left=45] node[above] {}
  (update.east);

  \path[->] (update.north) edge[bend left=45] node[above] {}
  (EID.south);
\end{tikzpicture}
\caption{Illustration of the necessary components for a general
  closed-loop, information-based sensing algorithm. Our primary
  contribution in this paper is using ergodic trajectory optimization
  for estimation (step C). We demonstrate implementation of closed-loop
  estimation for a particular sensing task using the SensorPod robot
  (Fig. \ref{FishB}), where the sensing task motivates choice of steps
  A, B, D. Section \ref{relatedwork} discusses alternative choices for
  steps A through C. \vspace{-10 pt} }\label{flow}
\end{figure}

Figure \ref{flow} illustrates the high level components for a general
estimation or mapping algorithm that iteratively collects sensor
measurements, updates an expected information map, and decides how to
acquire further measurements based on the information map. In this
section, we touch on related work for components A-C, although the
differentiating feature of the EEDI algorithm is the way in which
control decisions are made based on the expected information (step C in
Fig. \ref{flow}). The advantages of EEDI are discussed in Section
\ref{ergodicitydiscussion}.

\subsection{Representing the belief state}\label{relatedworkbelief}
The best choice for representing and updating the belief state for a
given application will depend on robot dynamics, sensor physics, and the
estimation task (modeling a field vs. target localization). Designing
appropriate representations for active sensing is a well-studied area of
research. For many applications, such as active sensing for
localization, parametric filters (e.g. the extended Kalman filter (EKF))
\cite{Sim05, Leun06a, Feder99, VanderHook2012} may be used. When the
posterior is not expected to be approximately Gaussian, nonparametric
filters, e.g. Bayesian filters \cite{Marchant2012, Wong05}, histogram
filters \cite{stachniss2003}, or particle filters \cite{kreucher2007,
  Roy2006, lu11} are often used. Mapping applications often use
occupancy grids \cite{Bourgault02i, Elfes89} or coverage maps
\cite{stachniss2003}, and much of the most recent work utilizes
  Gaussian processes to represent spatially varying phenomena or higher
  dimensional belief spaces, and the associated uncertainty
  \cite{Cao2013, Singh2009, Hoang2014, bender2013, low2008, souza2014}.
For the experimental work presented in this paper using the SensorPod
robot, we use a Bayesian filter as it is appropriate for general
(non-Gaussian) PDFs, sufficient for representing stationary estimation
objectives, and allows us to take into account sensor physics and
uncertainty in the estimate (see Section \ref{expsetup}). The
differentiating feature of the EEDI algorithm however---the ergodic
trajectory optimization---does not depend on the choice of belief
representation, so long as the choice enables calculation of an expected
information density map.

\subsection{ Calculating expected measurement utility}\label{measutcalc}
For a given sensing task and belief state, not all measurements are
equally informative. The quality of a measurement depends on the sensor
and may be distance, orientation, or motion dependent. To ensure useful
measurements are obtained given realistic time or energy restrictions,
sensing strategies for mobile sensors should seek to optimize
measurement quality \cite{Bajcsy88, spletzer03}. In some cases, it is
sufficient to consider only sensor field-of-view (i.e. useful
measurements can be obtained anywhere within a given distance from a
target), often called ``geometric sensing'' \cite{lu11, dasgupta2006,
  zhang09, hager91}. In many scenarios---if the search domain is
significantly larger than the sensor range---a geometric sensing
approach is sufficient. Many sensors, however, have sensitivity
characteristics within the range threshold that affect sensing efficacy.
Infrared range sensors, for example, have a maximum sensitivity region
\cite{Benet02}, and cameras have an optimal orientation and focal length
\cite{Denzler03}.

There are several different entropy-based measures from information
theory and optimal experiment design that can be used to predict
expected information gain prior to collecting measurements. Shannon
entropy has been used to measure uncertainty in estimation problems
\cite{Fox98, Arbel99, vazquez2001, takeuchi1998}, as well as
entropy-related metrics including Renyi Divergence\cite{Leun06a,
  kreucher05s}, mutual information, \cite{Toh06, Denzler02, zhang09,
  Tisd09a, Grocholsky06, Singh2009, Roy2006, lu2014}, entropy reduction
or information gain maximization \cite{zhang09B, hollinger2013}. In our
problem formulation we use Fisher information \cite{liao04, emery1998,
  Ucinski1999, Ucinski2000} to predict measurement utility. Often used
in maximum likelihood estimation, Fisher information quantifies the
ability of a random variable, in our case a measurement, to estimate an
unknown parameter \cite{Frie04, emery1998, liao04}. Fisher information
predicts that the locations where the ratio of the derivative of the
expected signal to the variance of the noise is high will give more
salient data (see Appendix \ref{Fisher Information} and the multimedia
attachment), and thus will be more useful for estimation.

In this paper, the Bayesian update mentioned in Section
\ref{relatedworkbelief} and the use of Fisher information to formulate
an information map are tools that allow us to close the loop on ergodic
control (update the map, step A in Fig. \ref{flow}), in a way that is
appropriate for the experimental platform and search objective (see
Appendix \ref{eediappendix}). The Bayesian update and the Fisher
information matter only in that they allow us to create a map of
expected information for the type of parameter estimation problems
presented in the examples in Section \ref{results}. Ergodic exploration
could, however, be performed over the expected information calculated
using different methods of representing the belief and expected
information, and for different applications such as those mentioned in
\ref{relatedworkbelief}.

\subsection{ Control for information acquisition }\label{measforcontrol}

In general, the problem of exhaustively searching for an optimally
informative solution over sensor state space and belief state is a
computationally intensive process, as it is necessary to calculate an
expectation over both the belief and the set of candidate control
actions \cite{Leun06a,Tisd09a,Singh2009,Atanasov2014}. Many algorithms
therefore rely on decomposing/discretizing the search space, the action
space, or both, and locally selecting the optimal sensing action
myopically (selecting only the optimal next configuration or control
input) \cite{kreucher05s, Feder99}. The expected information gain can,
for example, be locally optimized by selecting a control action based on
the gradient of the expected information \cite{Grocholsky06,
  kreucher2007, lu11, Bourgault02i}. As opposed to local information
maximization, a sensor can be controlled to move to that state which
maximizes the expected information globally over a bounded workspace
\cite{vazquez2001, Li05, liao04, Wong05, VanderHook2012}. Such global
information maximizing strategies are generally less sensitive to local
minima than local or gradient based strategies, but can result in
longer, less efficient trajectories when performed sequentially
\cite{Roy2006, stachniss2003}. While myopic information maximizing
strategies have an advantage in terms of computational tractability,
they are typically applied to situations where the sensor dynamics are
not considered \cite{vazquez2001, Li05, liao04, Wong05, VanderHook2012},
and even the global strategies are likely to suffer when uncertainty is
high and information diffuse (as argued in \cite{Rahimi2005,
  stachniss2003, low2008}, when discussing re-planning periods), as we
will see in Section \ref{results}.

To avoid sensitivity of single-step optimization methods to local
optima, methods of planning control actions over longer time
horizons---nonmyopic approaches---are often used. A great deal of
research in search strategies point out that the most general approach
to solving for nonmyopic control signals would involve solving a dynamic
programming problem \cite{Singh2009, Cao2013, low2008}, which is
generally computationally intensive. Instead, various heuristics are
used to approximate the dynamic programming solution \cite{Cao2013,
  low2008, Hoang2014, stachniss2003}. Variants of commonly used
sampling-based motion planners for maximizing the expected information
over a path for a mobile sensor have also been applied to sensor path
planning problems \cite{Cai2009, zhang09B, Hollinger2014, Sim05,
  Leun06a, Ryan2010}.
 
Search-based approaches are often not suitable for systems with dynamic
constraints; although they can be coupled with low-level (e.g. feedback
control) planners \cite{MartinezCantin2009,lu2014}, or dynamics can be
encoded into the cost of connecting nodes in a search graph
(``steering'' functions) \cite{Hollinger2014}, solutions are not
guaranteed to be optimal even in a local sense---both in terms of the
dynamics and the information---without introducing appropriate
heuristics \cite{Cao2013, low2008, Hoang2014, stachniss2003}. As we will
see in Section \ref{dynamics}, one of the advantages of EEDI is that it
naturally applies to general, nonlinear systems. We take advantage of
trajectory optimization techniques, locally solving for a solution to
the dynamic programming problem---assuming that the current state of the
system is approximately known.

Use of an ergodic metric for determining optimal control strategies was
originally presented in \cite{Mathew} for a nonuniform coverage problem.
The strategy in \cite{Mathew} involves discretizing the exploration time
and solving for the optimal control input at each time-step that
maximizes the rate of decrease of the ergodic metric. A similar method
is employed in \cite{Jacobs}, using a Mix Norm for coverage on
Riemannian manifolds. While our objective function includes the same
metric as \cite {Mathew}, the optimal control problem and applications
are different, notably in that we compute the ergodic trajectory for the
entire planning period $T$, and apply it to a changing belief state.
Additionally, the feedback controllers derived in \cite{Mathew} are
specific to linear, first- or second-order integrator systems, whereas
our method applies to general, nonlinear dynamic systems.

\section{Ergodic optimal control}
\label{ergodicitydiscussion}
\begin{figure}
  \centering \subfloat [Ergodic trajectory ]
  {\label{ergodiccartoon}\includegraphics[width=.7\columnwidth
    ]{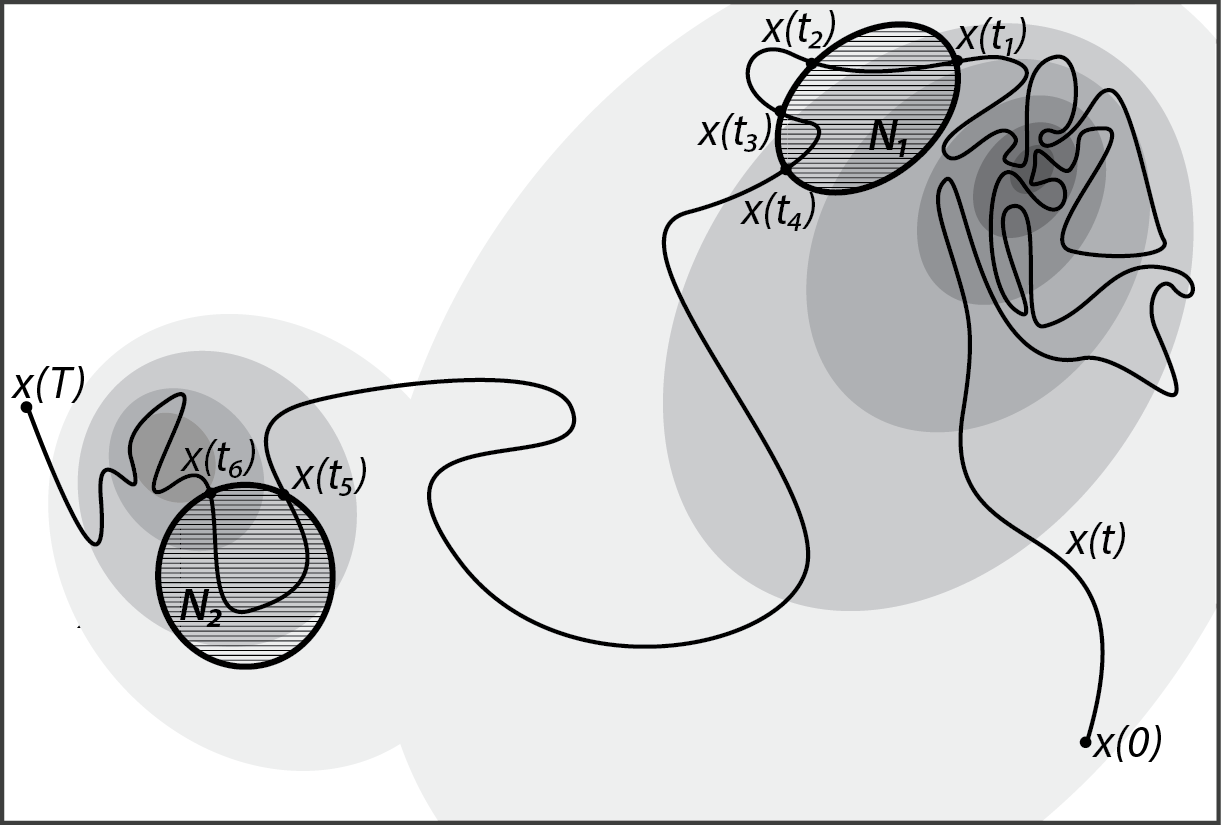}}\vfill

\subfloat
  [Information maximizing trajectory
]
  {\label{infomaxcartoon}\includegraphics[width=.7\columnwidth ]{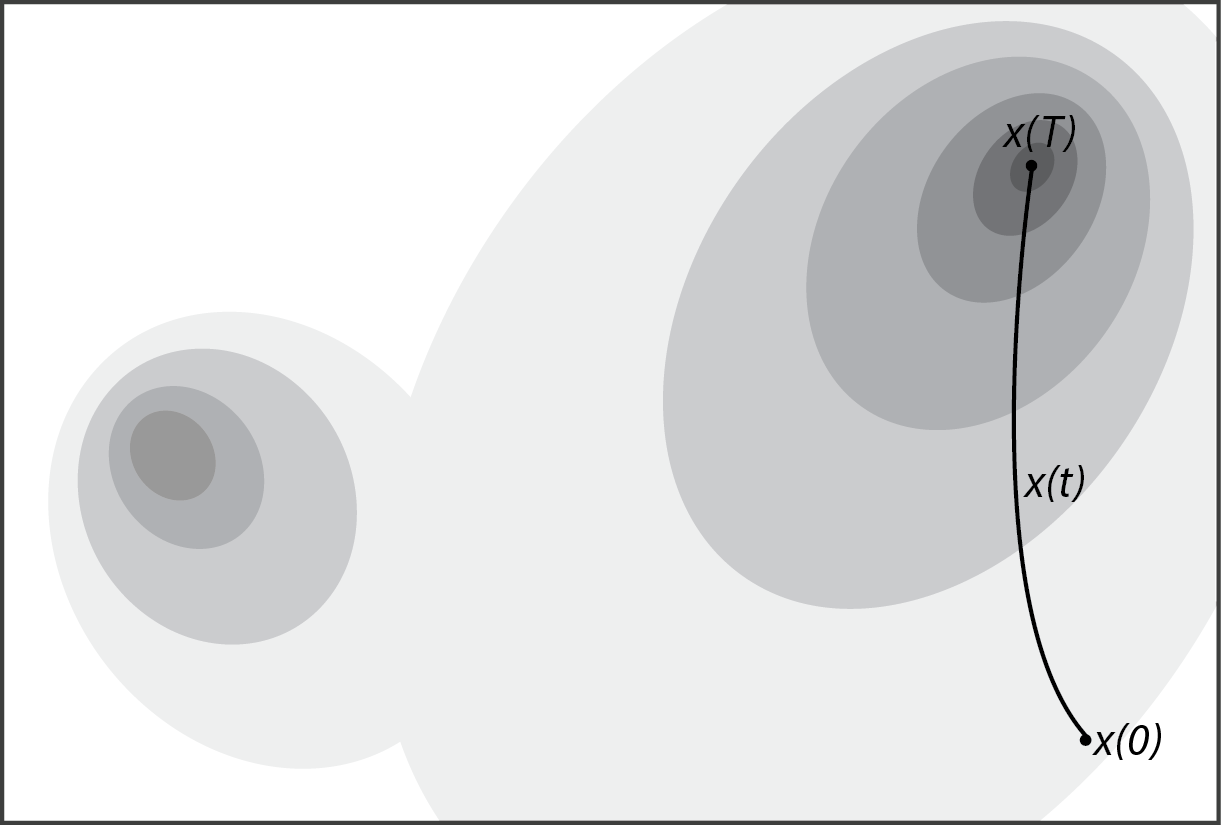}}
  \caption{Two candidate trajectories $x(t)$ for exploring the EID
    (depicted as level sets) are plotted in (a) and (b), both from $t=0$
    to $t=T$. Ergodic control provides a way of designing trajectories
    that spend time in areas proportional to how useful potential
    measurements are likely to be in those areas (a). This is in
    contrast to many alternative algorithms, which directly maximize
    integrated information gain over the trajectory based on the
    current best estimate, as in \ref{infomaxcartoon}. As illustrated in
    \ref{ergodiccartoon}, A trajectory $x(t)$ is \emph{ergodic} with
    respect to the PDF (level sets) if the percentage of time spent in
    any subset $N$ from $t=0$ to $t=T$ is equal to the measure of $N$;
    this condition must hold for all possible subsets. \vspace{-10 pt} }
\label{cartoon}
\end{figure}

Ergodic theory relates the time-averaged behavior of a system to the
space of all possible states of the system, and is primarily used in the
study of fluid mixing and communication. We use ergodicity to compare
the statistics of a search trajectory to a map of expected information
density (EID). The idea is that an efficient exploration strategy---the
path followed by a robot---should spend more time exploring regions of
space with higher expected information, where useful measurements are
most likely to be found. The robot should not, however, only visit the
highest information region (see Fig. \ref{infomaxcartoon}), but
distribute the amount of time spent searching proportional to the
overall EID (Fig. \ref{ergodiccartoon}).
 \label{infomaxcompare} This is the key distinction
between using ergodicity as an objective and previous work in active
sensing (e.g. information maximization); the ergodic metric encodes the
idea that, unless the expected information density is a delta function,
measurements should be \emph{distributed} among regions of high expected
information. Information maximizing strategies (that are also nonmyopic)
otherwise require heuristics in order to force subsequent measurements
away from previously sampled regions so as not to only sample the
information maxima.

As mentioned in Section \ref{relatedwork}, many commonly used algorithms
for active sensing, e.g. \cite{Feder99, Bourgault02i,
  Wong05,wilson2014}, involve a version of the type of behavior
illustrated in Fig. \ref{infomaxcartoon}, iteratively updating the EID
and maximizing information gain based on the current best estimate,
whether or not that estimate is correct. While computationally
efficient, globally information maximizing approaches are likely to fail
if the current estimate of the EID is wrong. In Section \ref{results},
for example, we show that even when the information map is updated while
calculating the information maximizing control, the estimate may get
trapped in a local maxima, e.g. when there is a distractor object that
is similar but not exactly the same as the target object.

Many sampling-based algorithms for information gathering therefore rely
on heuristics related to assuming submodularity between measurements,
e.g. assuming no additional information will be obtained from a point
once it has already been observed \cite{Singh2009, Sim05,
  Hollinger2014}. This assumption forces subsequent measurements away
from previously sampled regions so as not to only sample the information
maxima. As another way to distribute measurements, many nonmyopic
strategies select a set of waypoints based on the expected information,
and drive the system through those waypoints using search-based
algorithms \cite{zhang09B, zhang09, lu2014, dasgupta2006, Rahimi2005,
  souza2014, MartinezCantin2009}. Such approaches result in a predefined
sequence that may or may not be compatible with the system dynamics. If
the ordering of the waypoints is not predefined, target-based search
algorithms may require heuristics to avoid the combinatoric complexity
of a traveling salesman problem \cite{Song12, Kim2014}. In some cases,
search algorithms are not well-posed unless both an initial and final
(goal) position are specified \cite{zhang09, Cao2013}, which is not
generally the case when the objective is exploration.

Ergodic control enables how a robot searches a space to depend directly
on the dynamics, and is well posed for arbitrary dynamical systems. In
the case of nonlinear dynamics and nontrivial control synthesis,
encoding the search ergodically allows control synthesis to be solved
directly in terms of the metric, instead of in a hierarchy of problems
(starting with target selection and separately solving for the control
that acquires those targets, for example \cite{zhang09B,zhang09,
  dasgupta2006, Rahimi2005, souza2014, MartinezCantin2009}). In ergodic
trajectory optimization, the distribution of samples results from
introducing heuristics into the trajectory optimization, but of encoding
the statistics of the trajectory and the information map directly in the
objective. Using methods from optimal control, we directly calculate
trajectories that are ergodic with respect to a given information
density \cite{miller13R, miller13SE}. It is noteworthy, however, that
even if one wants to add waypoints to a search objective, ergodic search
is an effective means to drive the system to each waypoint in a
dynamically admissible manner (by replacing each waypoint with a
low-variance density function, thus avoiding the traveling salesman
problem). Further, ergodic control can be thought of as a way to
generate a continuum of dynamically compatible waypoints; it is similar
to \cite{zhang09B, zhang09, Rahimi2005}, but allows the number of
waypoints to go to $\infty$, making the control synthesis more tractable
for a broad array of systems.

 Many active sensing algorithms are formulated to
either prioritize exploitation (choosing locations based on the current
belief state) or exploration (choosing locations that reduce uncertainty
in the belief state); they are best suited for greedy, reactive
sampling, requiring a prior estimate \cite{Sim05}, or for coverage
\cite{Singh2009, Acar2003, choset2001}. Algorithms that balance both
exploration and exploitation typically involve encoding the two
objectives separately and switching between them based on some condition
on the estimate, \cite{Krause2007, low2008}, or defining a (potentially
arbitrary) weighted cost function that balances the tradeoff between the
two objectives \cite{Hoang2014, MartinezCantin2009, souza2014,
  marchant2014}. Using ergodicity as an objective results in an
algorithm that is suitable for both exploration-prioritizing coverage
sampling or exploitation-prioritizing ``hotspot'' sampling, without
modification (policy switching or user-defined weighted objectives
\cite{Krause2007, low2008, Hoang2014, MartinezCantin2009,
  marchant2014}). Moreover, the ergodic metric can be used in
combination with other metrics, like a tracking cost or a terminal cost,
but does not require either to be well-posed.

\subsection{Measuring Ergodicity}
We use the \emph{distance from ergodicity} between the time-averaged
trajectory and the expected information density as a metric to be
minimized. We assume a bounded, $n$-dimensional workspace (the search
domain) $X \subset \real^{n}$ defined as
$[0,L_1]\times[0,L_2]...\times[0,L_n]$. We define $\bm x(t)$ as the
sensor trajectory in workspace coordinates, and the density function
representing the expected information density as $EID(\bm x)$.

The spatial statistics of a trajectory $\bm x(t)$ are quantified by the
percentage of time spent in each region of the workspace,
\begin{equation}\label{timeavedist}C(\bm x)=\frac{1}{T}\int_0^T \delta\left[\bm x-\bm x(t))\right]dt,
\end{equation} 
where $\delta$ is the Dirac delta \cite{Mathew}. The goal is to drive
the spatial statistics of a trajectory $\bm x(t)$ to match those of the
distribution $EID(\bm x)$; this requires choice of a norm on the
difference between the distributions $EID(\bm x)$ and $C(\bm x)$. We
quantify the difference between the distributions, i.e. the distance
from ergodicity, using the sum of the weighted squared distance between
the Fourier coefficients $\phi_{\bm k}$ of the EID, and the coefficients
$c_k$ of distribution representing the time-averaged
trajectory.\footnote{The Fourier coefficients
    $\phi_{\bm k}$ of the distribution $\Phi(\bm x)$ are computed using
    an inner product,
    $\phi_{\boldsymbol k}=\int_{X}\phi(\bm x)F_{\boldsymbol k}(\bm x)d
    \bm x,$
    and the Fourier coefficients of the basis functions along a
    trajectory $\bm x(t)$, averaged over time, are calculated as
    $\label{ck2} c_{\boldsymbol k}(\bm
    x(t))=\frac{1}{T}\int_{0}^{T}F_{\boldsymbol k}(\bm x(t))dt, $
    where $T$ is the final time and $F_k$ is a Fourier basis function.}
The ergodic metric will be defined as $\mathcal{E}$, as follows:
\begin{equation}\label{ephi}
  \mathcal{E}(\bm x(t))=\sum_{\boldsymbol k=0 \in \mathbb{Z}^n}^{\boldsymbol K \in \mathbb{Z}^n}\Lambda_k\left[c_{\boldsymbol k}(\bm x(t))-\phi_{\boldsymbol k}\right]^2 
\end{equation}
where $\boldsymbol K$ is the number of coefficients calculated along each of the $n$
dimensions, and $\boldsymbol {k}$ is  a multi-index 
$(k_1,k_2,...,k_n)$. Following \cite{Mathew},
$\Lambda_{\boldsymbol k} = \frac{1}{(1+||{\boldsymbol k}||^2)^s}$ is a
weight where $s=\frac{n+1}{2}$, which places larger weight  on lower
frequency information.

Note that the notion of ergodicity used here does not strictly require
the use of Fourier coefficients in constructing an objective function.
The primary motivation in using the norm on the Fourier coefficients to
formulate the ergodic objective is that it provides a metric that is
differentiable with respect to the trajectory $\bm x(t)$. This
particular formulation is not essential---any differentiable method of
comparing the statistics of a desired expected information density to
the spatial distribution generated by a trajectory will suffice, however
finding such a method is nontrivial. The Kullback-Leibler (KL)
divergence or Jensen-Shannon (JS) divergence \cite{bender2013}, for
example, commonly used metrics on the distance between two
distributions, are not differentiable with respect to the trajectory
$\bm x(t)$.\footnote{Due to the Dirac delta in Eq. \eqref{timeavedist},
the JS divergence ends up involving evaluating the EID along the
trajectory $\bm x(t)$. In general we do not expect to have a closed form
expression for the EID, so this metric is not differentiable in a way
that permits trajectory optimization. Alternatively, replacing the Dirac
delta in Eq. \eqref{timeavedist} with a differentiable approximation
(e.g. a Gaussian) would expand the range of metrics on ergodicity, but
would introduce additional computational expense of evaluating an $N$
dimensional integral when calculating the metric and its derivative.} On
the other hand, by first decomposing both distributions into their
Fourier coefficients, the inner product between the transform and the
expression for the time-averaged distribution results in an objective
that is differentiable with respect to the trajectory.

\subsection{Trajectory Optimization} \label{trajopt}

For a general, deterministic, dynamic model for a mobile sensor
$ \dot{\bm x}(t)=f(\bm x(t),\bm u(t)) $, where $\bm x\in\mathbb{R}^N$ is
the state and $\bm u\in\mathbb{R}^n$ the control, we can solve for a
continuous trajectory that minimizes an objective function based on both
the measure of the ergodicity of the trajectory with respect to the EID
and the control effort, defined as
\begin{equation}\label{Jdis2} 
  J(\bm x(t))=\underbrace{\gamma \mathcal{E}[\bm x(t)]
}_\text{ergodic  cost} +\underbrace{\int_{0}^{T}\frac{1}{2}
  \bm u(\tau)^TR  \bm u(\tau)d\tau}_\text{control effort}.
\end{equation} 
In this equation, $\gamma \in \mathbb{R}$ and
$R(\tau)\in \mathbb{R}^{m\times m}$ are arbitrary design parameters that
affect the relative importance of minimizing the distance from
ergodicity and the integrated control effort. The choice of ratio
of $\gamma$ to $R$ plays the exact same role in ergodic control as it
does in linear quadratic control and other methods of optimal control;
the ratio determines the balance between the objective---in this case
ergodicity---and the control cost of that objective. Just as in these
other methods, changing the ratio will lead to trajectories that perform
better or worse with either more or less control cost.

In \cite{miller13R} we show that minimization of Eq. \eqref{Jdis2} can
be accomplished using an extension of trajectory optimization
\cite{Hauser}, and derive the necessary conditions for optimality. The
extension of the projection-based trajectory optimization method from
\cite{Hauser} is not trivial as the ergodic metric is not a Bolza
problem; however, \cite{miller13R} proves that the first-order
approximation of minimizing Eq. \eqref{Jdis2} subject to the dynamics
$ \dot{\bm x}(t)=f(\bm x(t),\bm u(t))$ is a Bolza problem and that
trajectory optimization can be applied to the ergodic objective. The
optimization does not require discretization of search space or control
actions in space or time. While the long time horizon optimization we
use is more computationally expensive than the myopic, gradient-based
approach in \cite{Mathew}, each iteration of the optimization involves a
calculation with known complexity. The EID map and ergodic objective
function could, however, also be utilized within an alternative
trajectory optimization framework (e.g. using sequential quadratic
programming). Additionally, for the special case of
$\dot{\bm x} = \bm u$, sample-based algorithms \cite{Hollinger2014} may
be able to produce locally optimal ergodic trajectories that are
equivalent (in ergodicity) to the solution obtained using trajectory
optimization methods; this would not however, be the case for general,
nonlinear dynamics $ \dot{\bm x}(t)=f(\bm x(t),\bm u(t))$.

Ergodic optimal control allows for the time of exploration to be
considered as an explicit design variable. It can, of course, be of
short duration or long duration, but our motivation is largely long
duration. The idea is that one may want to execute a long exploration
trajectory prior to re-planning. The most straightforward motivation is
that integrating measurements and updating the belief may be the more
computationally expensive part of the search algorithm \cite{Roy2006,
  stachniss2003, Sim05, low2008}. Overly reactive/adaptive
strategies---strategies that incorporate measurements as they are
received---are also likely to perform poorly when the estimate
uncertainty is high \cite{Rahimi2005, stachniss2003,low2008} or in the
presence of (inevitable) modeling error. If, for example, the
measurement uncertainty is not perfectly captured by the measurement
model, the idealized Bayesian update can lead to overly reactive control
responses. Instead, one may wish to take enough data such that the
central limit theorem can be applied to the measurement model, so that
the measurement model is only anticipated to be applicable on average
over the length of the exploratory motion \cite{Chirikjian2009}. Future
work will involve exploring the effects of reducing the re-planning
horizon on the success of the estimation algorithm.

\subsection{Assumptions: Ergodic Trajectory Optimization}\label{ergoassumptions}
Ergodic trajectory optimization requires a controllable motion model for
the robot, and an expected information density function defined over the
sensor state space. The motion model can be nonlinear and/or dynamic,
one of the primary benefits of a trajectory optimization approach. For
this paper we consider calculating search trajectories in one and two
dimensions (although the sensor dynamics can be higher dimensional). The
trajectory optimization method can be extended to search in higher
dimensional search spaces such as $\real^3$ and $SE(2)$, so long as a
Fourier transform \cite{Chirikjian} exists for the manifold
\cite{miller13SE}. We consider only uncertainty of static environmental
parameters (e.g. fixed location and radius of an external target)
assuming noisy measurements. We assume deterministic dynamics.

 \section{Experimental methods: search for stationary targets using the
   Sensorpod Robot }
 \label{expsetup}

 Although ergodic trajectory optimization is general to sensing
 objectives with spatially distributed information, we describe an
 application where the belief representation and expected information
 density (EID) calculation (steps A, B, and D in Fig. \ref{flow}) are
 chosen for active localization of stationary targets using the
 SensorPod robot. This allows us to experimentally test and validate a
 closed-loop version of the ergodic optimal control algorithm, described
 in Section \ref{ergodicitydiscussion}, against several established
 alternatives for planning control algorithms based on an information
 map.

\begin{figure}[!t] 
   \centering 
 \vspace{-0pt}
 \subfloat[+0.2 mV  expected voltage difference between the sensors  on the SensorPod for a target located as shown.] {\label{isopotentialsb}\includegraphics[trim=-.3in 0in -.3in
    .0in,clip=true,width=1\columnwidth ]{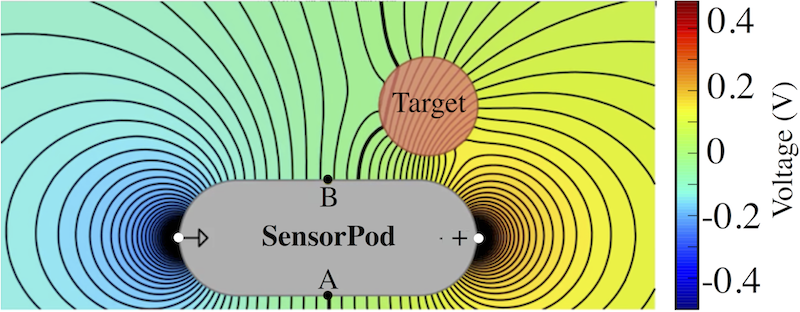}}\\
 \subfloat[-0.2 mV  expected voltage difference between the sensors (A-B)  on the SensorPod for a target located as shown.] {\label{isopotentialsa}\includegraphics[trim=-.3in 0in -.3in
    0in,clip=true,width=1\columnwidth ]{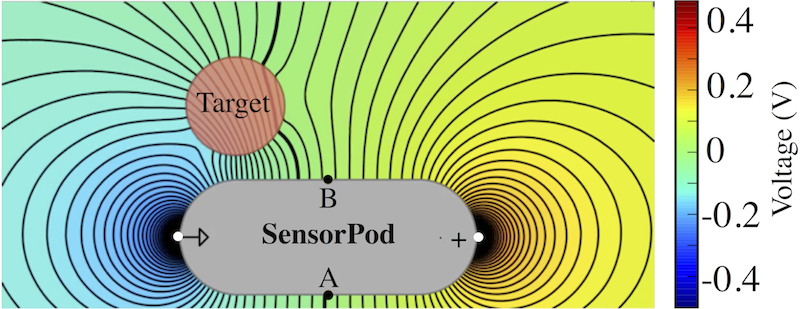}}
  \caption{The SensorPod (grey) measures the difference between the
    field voltage at the two sensors (A-B). The field is generated by
    two excitation electrodes on the SensorPod body. The field
    (simulated isopotential lines plotted in black) changes in the
    presence of a target with different conductivity than the
    surrounding water. The 0V line is bolded. The perturbation cause by
    the same object results in a different differential measurement
    between sensors A and B based on the position of the object relative
    to the SensorPod. For more information and an animation of this
    plot, please see the multimedia video attachment. Note that the
    SensorPod is not measuring the field itself (which is emitted by,
    and moves with, the robot), but the voltage differential between two
    sensors induced by disturbances in the field. }
 \label{isopotentials} 
 \vspace{-10pt}
 \end{figure}

 \begin{figure}[!t] \centering \vspace{-0 pt} \subfloat[The expected
   differential voltage measurement (A-B from Fig. \ref{isopotentials}) is plotted as a function of robot
   centroid for a target (pink) located at X,Y=(0,0). Two possible
   SensorPod trajectories are plotted in black (solid and dashed). The  target is placed below the robot's plane of motion to prevent collisions.]
   {\label{Fishmeasa}\includegraphics[trim=-1in .02in -1in
     .0in,clip=true,width=\columnwidth ]{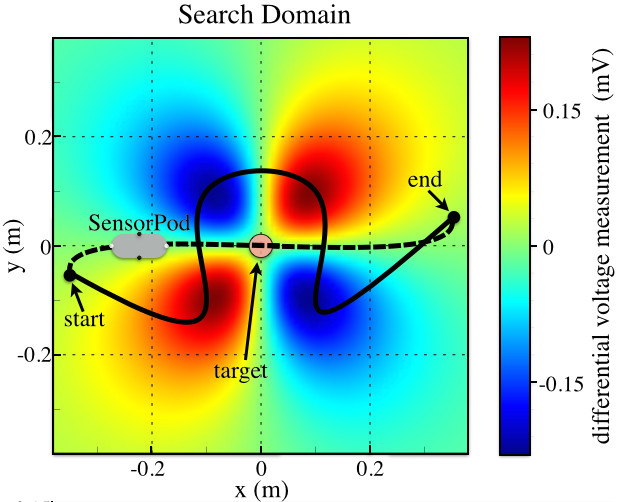}}\\
   \subfloat[Simulated differential voltage measurements for the trajectories in
   (a) are plotted as a function of time. ]
   {\label{Fishmeasb}\includegraphics[trim=-1in .02in -1in
     .025in,clip=true,width= \columnwidth ]{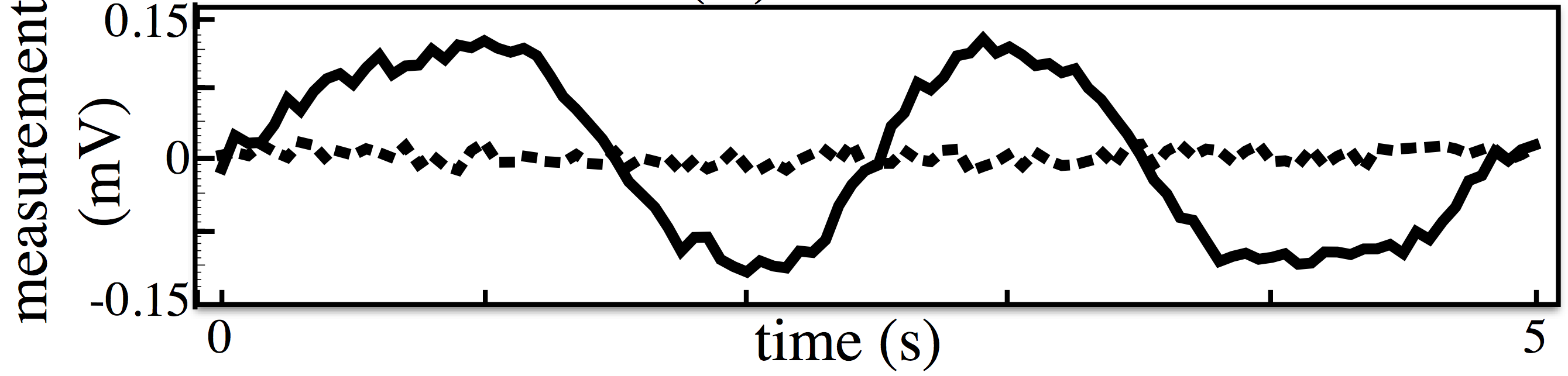}}
  \caption{
    Measurements collected by the SensorPod have a nonlinear
and non-unique mapping to target
location. The dashed trajectory in Fig. \ref{Fishmeasa} 
    yields  uninformative measurements (possible to observe for many potential target locations); the solid trajectory in Fig.
    \ref{Fishmeasa}, produces a series of measurements that are unique
    for that target position, and therefore useful for estimation. }
  \label{Fishmeas} \vspace{-10pt} 
\end{figure}

Inspired by the electric fish, the SensorPod (Fig. \ref{FishB}) has two
excitation electrodes that create an oscillating electric field. We use
a single pair of voltage sensors---hence, a \emph{one-dimensional
  signal} ---on the body of the SensorPod to detect perturbations in the
field due to the presence of underwater, stationary, nonconducting
spherical targets. The expected measurement depends on the location,
size, shape, and conductivity of an object as well as the strength of
the electric field generated by the robot; for details, see
\cite{Bai2015}. The perturbed electric fields and resulting differential
measurements for targets in two locations relative to the SensorPod are
shown in Fig. \ref{isopotentials}, and the differential voltage
measurement is plotted in Fig. \ref{Fishmeasa}. Figure \ref{Fishmeasb}
shows the expected differential measurement for two candidate sensor
trajectories. The multimedia attachment provides additional intuition
regarding the SensorPod and and the observation model. The solid line
trajectory is more informative, as measured using Fisher Information,
than the dashed line; our goal is to automatically synthesize
trajectories that are similarly more informative.

The objective in the experimental results presented in Section
\ref{results} is to estimate a set of unknown, static, parameters
describing individual spherical underwater targets. Details and
assumptions for implementation of both the Bayesian filter and the
calculation of the expected information density for the SensorPod robot,
including for the multiple target case, can be found in Appendix
\ref{eediappendix}; an overview of the algorithm is provided here,
corresponding to the diagram in Fig. \ref{flow}). For a graphical,
animated overview of the algorithm, please also see the attached
multimedia video.

\begin{figure}[!t] \centering \vspace{-10 pt} 
  \subfloat[A low-variance PDF of 2D target location ]
  {\label{tightprob}\includegraphics[trim=.1in .02in .27in
    .025in,clip=true,width=.31 \columnwidth ]{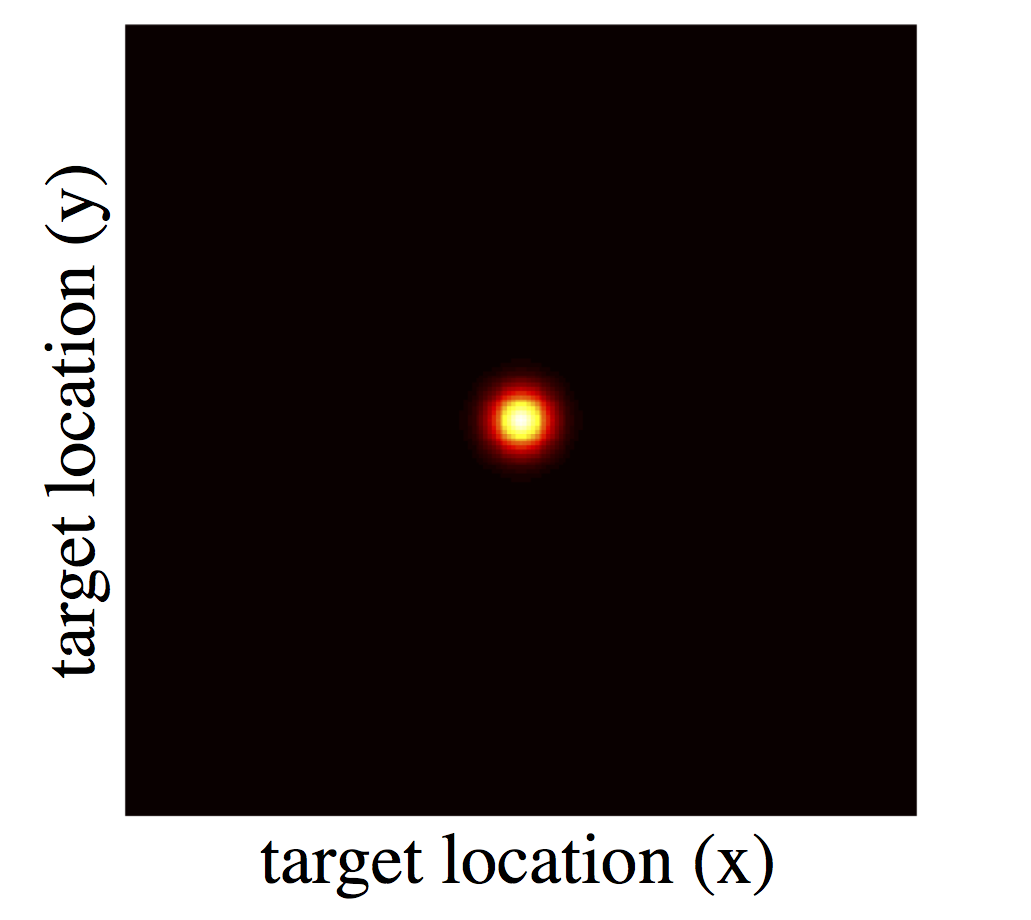}}
\hspace{2 pt}
\subfloat[ EID  for  target location for the PDF in (a)]
  {\label{FItight}\includegraphics[trim=-.0in .02in .4in
    .025in,clip=true,width=.31 \columnwidth
    ]{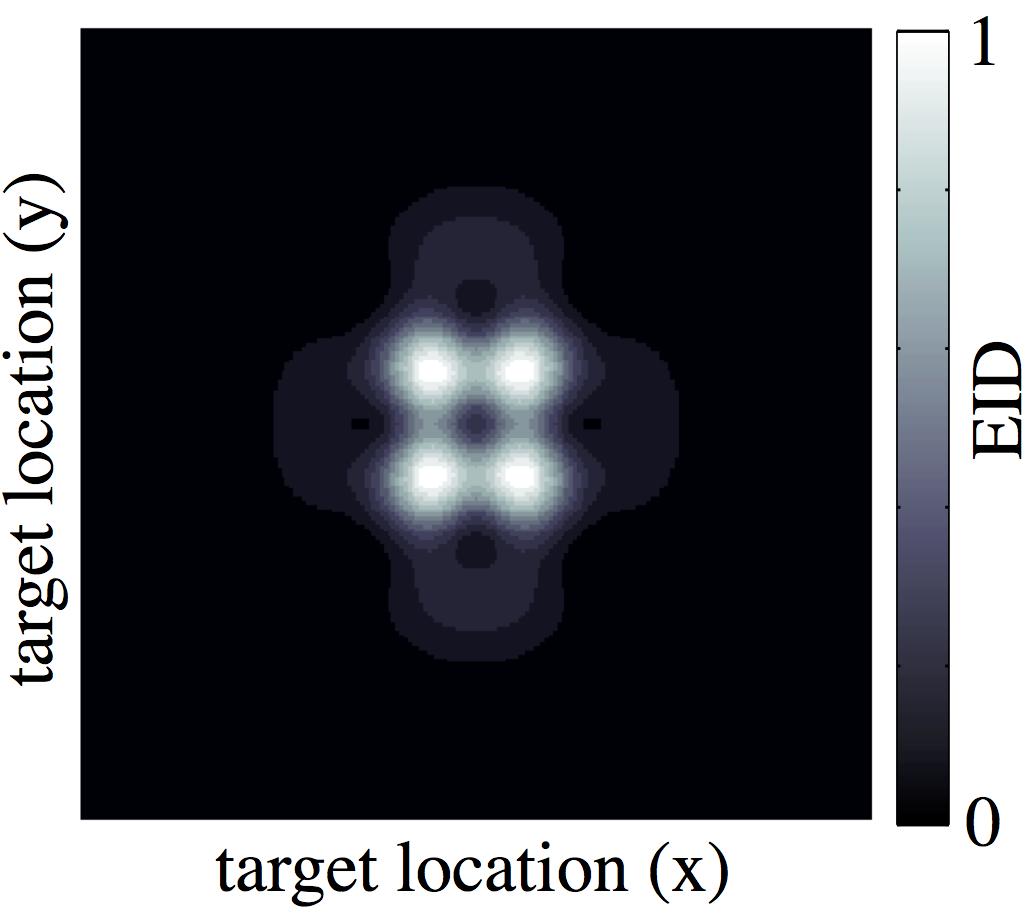}}
\hspace{2 pt}
\subfloat[EID for  target
  location and radius.  ]{\label{FIrad}
\includegraphics[trim=-0.in .02in .4in
    .065in,clip=true,width=.31\columnwidth ]{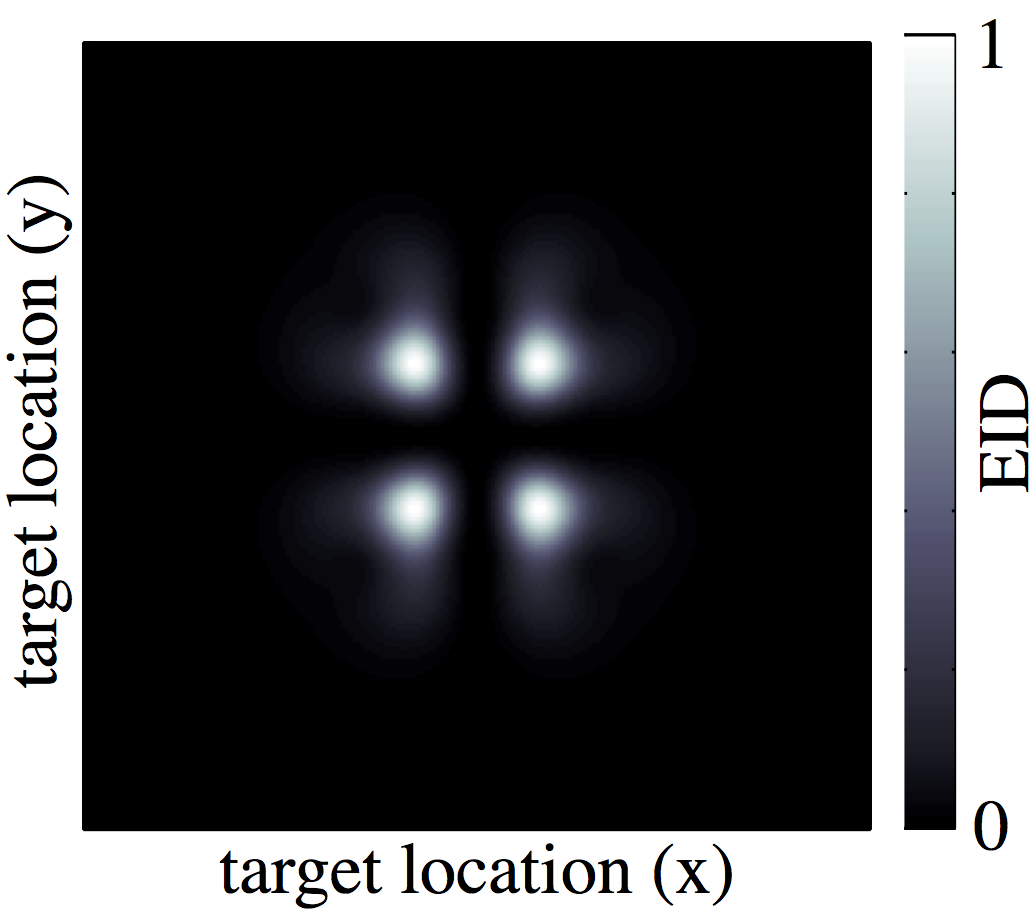}}
  \vspace{-5 pt}\\
  \subfloat[ A higher-variance PDF of  2D target location.] {\label{smearprob}\includegraphics[trim=-.1in
    .02in .3in .025in,clip=true,width=.34 \columnwidth
    ]{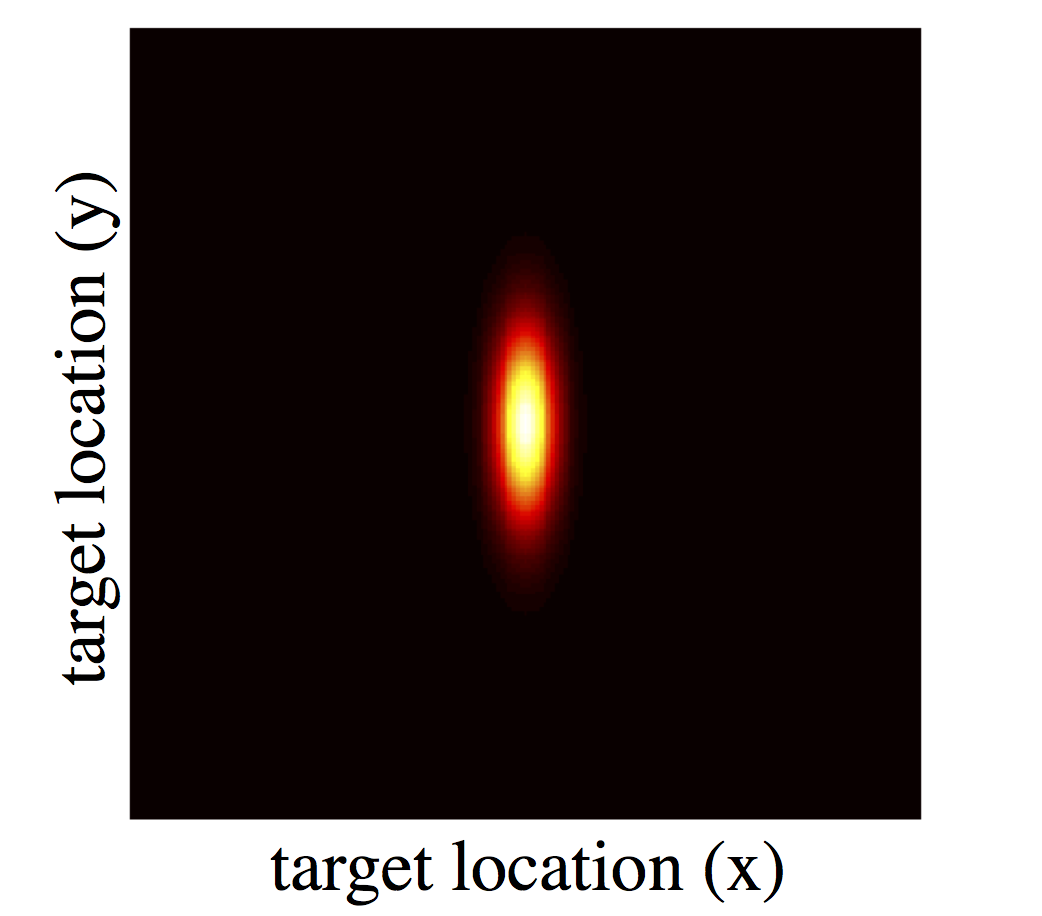}} \quad 
\subfloat[ EID map for the PDF in
  (c) ]{\label{FIsmear}\includegraphics[trim=-.1in .02in .37in
    .025in,clip=true,width=0.33\columnwidth ]{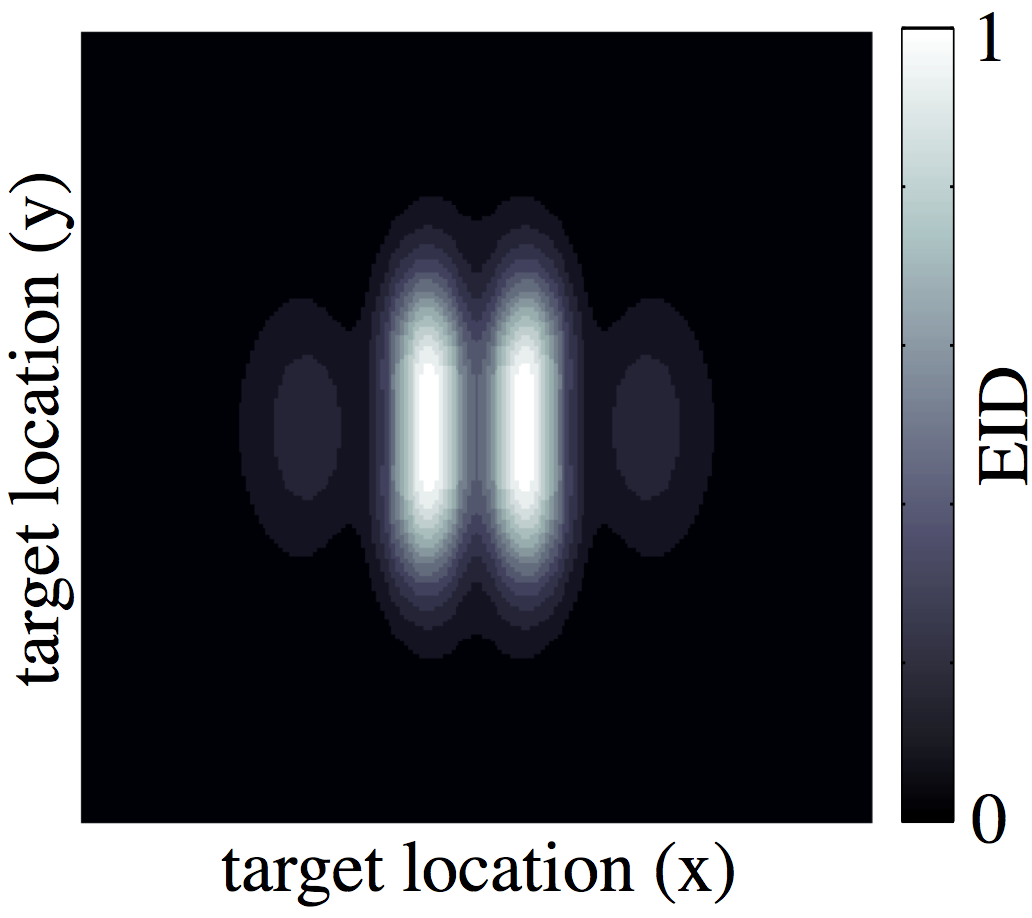}
  }\vspace{-0 pt}
  \caption{
The EID is dependent on the measurement model and the current
probabilistic estimate. 
Figures \ref{FItight}, \ref{FIrad}, \ref{FIsmear} show examples of the
EID for different PDFs and estimation tasks for the SensorPod
measurement model. For \ref{FIrad}, the projection of the corresponding PDF (defined in three-dimensions) onto the 2D
  location space would be  similar to (a). The EID is calculated according to Eq. \eqref{eideq}.
In all cases, calculation of the EID produces a map over the search
domain, regardless of the estimation task. 
}
  \label{FIsmaps} \vspace{-10pt} 
\end{figure}

The algorithm is initialized with the sensor state at the
initial time $\bm x(0)$ and an initial probability distribution
$p(\bm \theta)$ for the parameters $\bm\theta$. We represent and update
the parameter estimate using a Bayesian filter, which updates the
estimated belief based on collected measurements (Fig. \ref{flow}, step
A). The initial distribution can be chosen based on prior information
or, in the case of no prior knowledge, assumed to be uniform on
bounded domains. At every iteration of the EEDI algorithm, the EID is
calculated by taking the expected value of the Fisher information with
respect to the belief $p(\bm \theta)$ (Fig. \ref{flow}, step B). For
estimation of multiple parameters, we use the D-optimality metric on the
expected Fisher information matrix, equivalent to maximizing the
determinant of the expected information \cite{emery1998}.\footnote{Note
  that alternative choices of optimality criteria may result in
  different performance for different problems based on, for example,
  the conditioning of the information matrix. D-optimality is commonly
  used for similar applications and we found it to work well
  experimentally; however the rest of the EEDI algorithm is not
  dependent on this choice of optimality criterion.} In Fig.
\ref{FIsmaps}, the corresponding EIDs for two different belief maps for
2D target location (Figs. \ref{FItight} and \ref{FIsmear}), as well as
the EID for estimating both 2D target location and target radius
(\ref{FIrad}), are shown. The EID is always defined over the sensor
configuration space (2D), although the belief map may be in a different
or higher dimensional space (e.g. over the 2D workspace and the space of
potential target radii). The normalized EID is used to calculate an
optimally ergodic search trajectory for a finite time horizon (Fig.
\ref{flow}, step C). The trajectory is then executed, collecting a
series of measurements (Fig. \ref{flow}, step D, for time $T$).
Measurements collected in step D are then used to update the belief
$p(\bm \theta)$, which is then used to calculate the EID in the next
EEDI iteration. The algorithm terminates when the norm on the of the
estimate falls below a specified value.

For localizing and estimating parameters for multiple targets, we
initialize the estimation algorithm by assuming that there is a single
target present, and only when the norm on the variance of the parameters
describing that target fall below the tolerance $\epsilon$ do we
introduce a second target into the estimation. The algorithm stops
searching for new targets when one of two things happen: 1) parameters
for the last target added converge to values that match those describing
a target previously estimated (this would only happen if all targets
have been found, as the EID takes into account expected measurements
from previous targets), or 2) parameters converge to an invalid value
(e.g. a location outside of the search domain), indicating failure. The
algorithm terminates when the entropy of the belief map for all targets
falls below a chosen value; for the 0 target case, this means that the
SensorPod has determined that there are no objects within the search
space. Note that the EID for new targets takes into account the
previously located targets

\subsection{Assumptions: stationary target localization using the SensorPod (EEDI example)}\label{expassumptions}
We make a number of assumptions in choosing steps A, B, and D in Fig.
\ref{flow}, detailed in Appendix \ref{eediappendix}. We assume a
measurement is made according to a known, differentiable measurement
model (a function of sensor location and parameters), and assume the
measurements have zero-mean, Gaussian, additive
noise.\footnote{Related work in active
    electrosense has shown that zero mean Gaussian is a reasonable
    assumption for sensor noise \cite{Solb08a}.} We assume independence
between individual measurements, given that the SensorPod state is known
and the measurement model is not time-varying. Measurement independence
is commonly assumed, for example in occupancy grid problems
\cite{stachniss2003}, however more sophisticated likelihood functions
that do not rely on this assumption of independence \cite{Thrun2003}
could be used without significantly changing the structure of the
algorithm.

\begin{figure*}[!t] \vspace{-10pt} \centering \subfloat[For estimation of
  target location in 1D (Sections \ref{POC1D} and \ref{IC}), the target
  object (green) was placed at a fixed distance of $y= 0.2$ m from the
  SensorPod line of motion, and the distractor (pink) at $y_d= 0.25$ m.
  ] {\label{1dtanks} \includegraphics[width=.62 \columnwidth
    ]{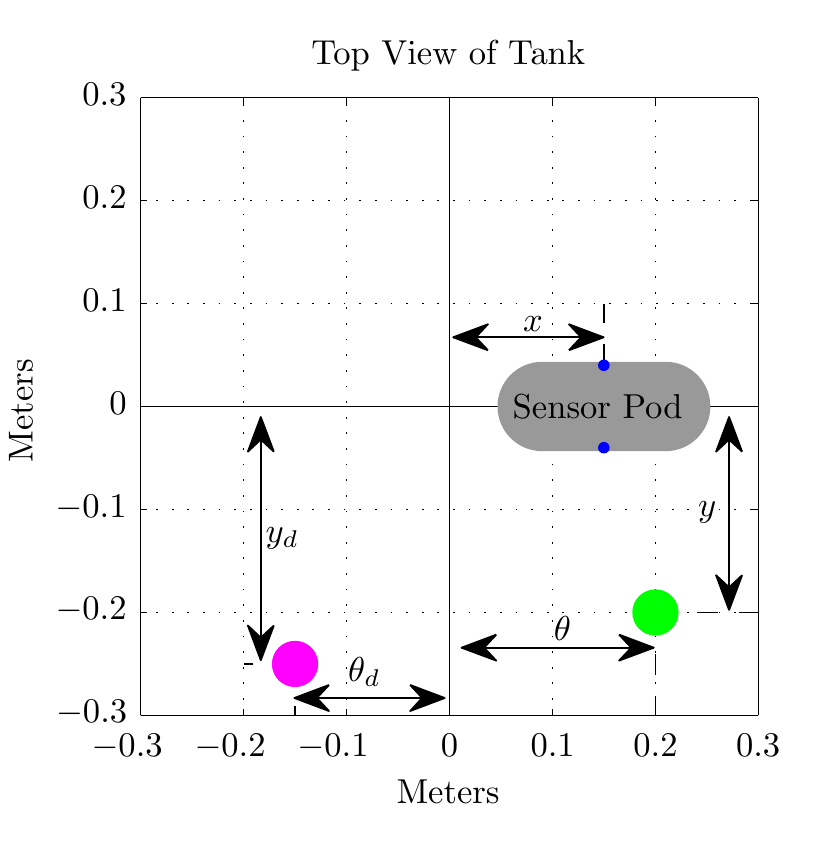}} \quad \subfloat[ Expected voltage
  measurement over 1D sensor state for the target (pink) and distractor
  (green) objects alone, and when both target and distractor are present
  (black). ] {\label{fig:voltagetrace} \includegraphics[width=.7
    \columnwidth ]{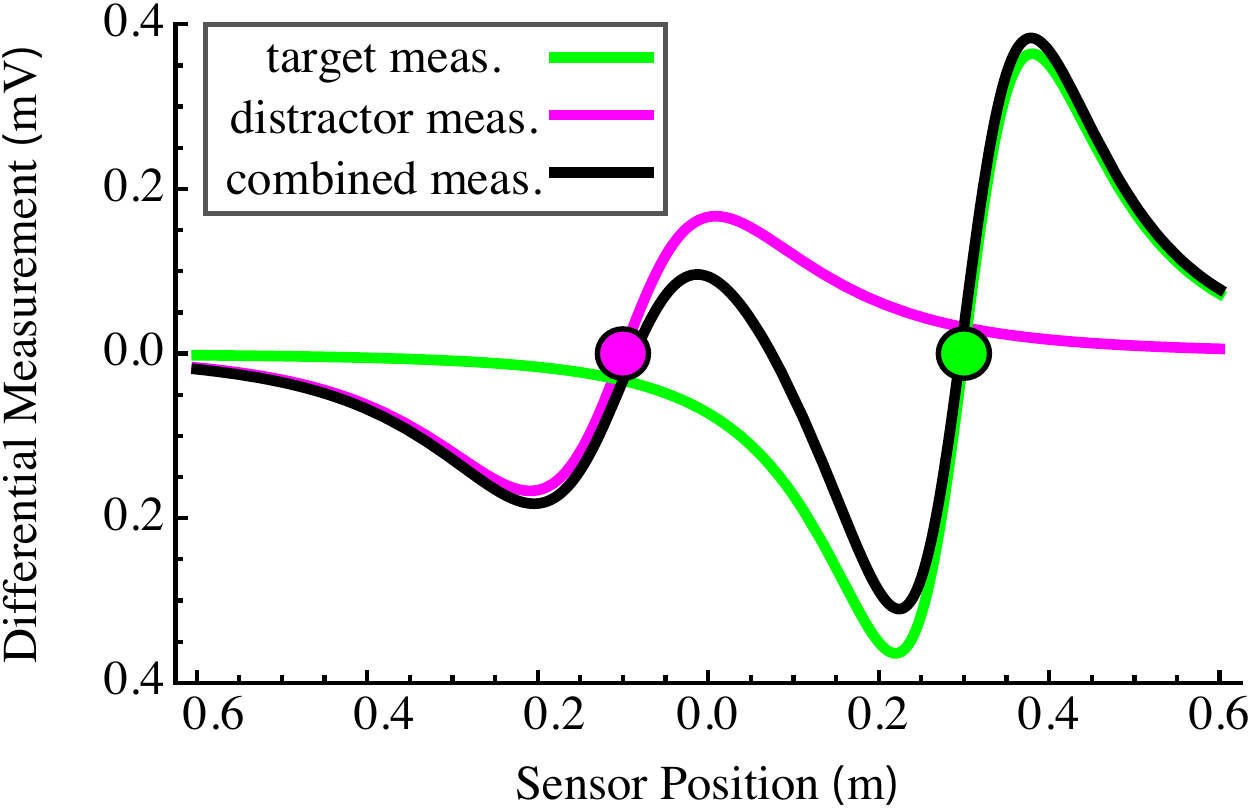}}\quad \subfloat[Example of
  tank configuration for 2D localization of two targets. For all trials,
  SensorPod and object locations are measured from the center of the
  tank.\vspace{-0 in }] {\label{2dtanks}\includegraphics[width=.62
    \columnwidth ]{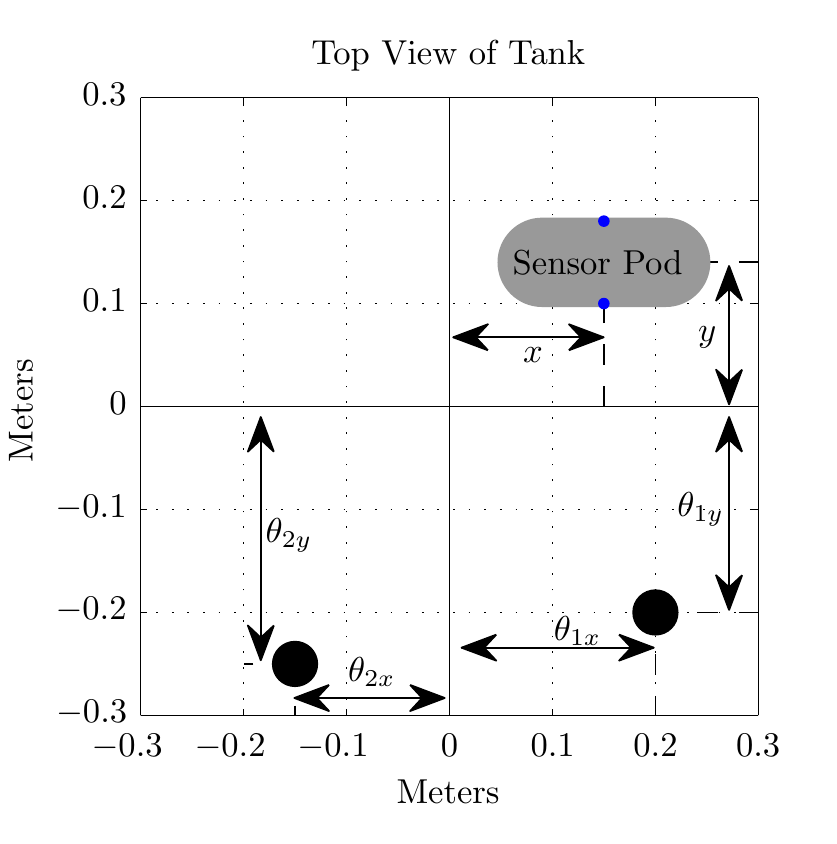}}
 \caption{
   Targets were placed below the robot's plane of motion to prevent
   collisions. The orientation of the robot is held constant. The
   voltage sensors sample at 100 Hz, with an assumed standard deviation
   of $100$ $\mu V$ for the measurement noise, the experimentally
   observed noise level of the SensorPod sensors. \vspace{-10 pt}}
 \label{1dtraj1} 
 \vspace{-0cm}
 \end{figure*}

 For the single target case, we maintain a joint probability
 distribution for parameters describing the same target as they are
 likely to be highly correlated. In order to make the problem of finding
 an arbitrary number of objects tractable, we assume that the parameters
 describing different targets are independent and that a general
 additive measurement model may be used, similar to \cite{Leun06a,
   thrun05p, Wong05}. Although the voltage perturbations from multiple
 objects in an electric field do not generally add linearly, we make the
 assumption that the expected measurement for multiple objects can be
 approximated by summing the expected measurement for individual
 objects, which simplifies calculations in Appendix
 \ref{eediappendix}.\footnote{Additional experimental work (not
     shown), demonstrated that at a minimum of 6 cm separation between
     objects, there is no measurable error using this approximation; in
     the experimental and simulated trials, we use a minimum separation
     of 12 cm.} While the computation of Fisher Information and the
 likelihood function used for the Bayesian update depend on the
 assumptions mentioned above, the ergodic optimal control calculations
 do not, and only depend on the existence of an EID map.

 The SensorPod is attached to a 4-DOF gantry system, which allows use of
 a kinematic model of the SensorPod in Eq. \eqref{Jdis2}, i.e. the
 equations of motion for the experimental system are
 $\dot{\bm x}(t)=\bm u(t)$, where $\bm x$ is the SensorPod position in
 1D (Sections \ref{POC1D} and \ref{IC}) or 2D (Sections \ref{POC2D},
 \ref {POCmultiple}, and \ref{2dtargetradius}). The kinematic model and
 2D search space also enable comparison with other search methods;
 however, it should be noted that EEDI is applicable to dynamic,
 nonlinear systems as well, as will be demonstrated in simulation in
 Section \ref{dynamics}.

 Ergodic trajectory optimization, presented in Section
 \ref{ergodicitydiscussion}, calculates a trajectory for a fixed-length
 time horizon $T$, assuming that the belief, and therefore the EID map,
 remains stationary over the course of that time horizon. In the
 following experiments, this means that each iteration of the closed
 loop algorithm illustrated in Fig. \ref{flow} involves calculating a
 trajectory for a fixed time horizon, executing that trajectory in its
 entirety, and using the series of collected measurements to update the
 EID map before calculating the subsequent trajectory. The complete
 search trajectory, from initialization until termination, is therefore
 comprised of a series of individual trajectories of length $T$, where
 the belief and EID are updated in between (this is also true for the
 alternative strategies used for comparison in Section \ref{results}).
 The EID map could alternatively be updated and the ergodic trajectory
 re-planned after each measurement or subset of measurements, in a more
 traditional receding horizon fashion, or the time horizon (for planning
 and updating) could be optimized.

\subsection{Performance assessment}
In the experiments in Section \ref{results}, we assess performance using
\textbf{\emph{time to completion}} and \textbf{\emph{success rate}}.
Time to completion refers to the time elapsed before the termination
criterion is reached, and a successful estimate obtained. We present
results for time until completion as the “slowdown factor.” The
\textbf{\emph{slowdown factor}} is a normalization based on minimum time
until completion for a particular set of experiments or simulation. For
a trial to be considered successful, the mean of the estimate must be
within a specified range of the true target location, and in Section
\ref{POCmultiple}, the number of targets found must be correct. The
tolerance used on the distance of the estimated parameter mean to the
true parameter values was 1 cm for the 1D estimation experiments and 2
cm for 2D experiments. In both cases this distance was more than twice
the standard deviation used for our termination criterion.

A maximum run-time was enforced in all cases (100 seconds for 1D and
1000 seconds for 2D experiments). For simple experimental scenarios,
e.g. estimation of the location of a single target in 2D (Section
\ref{POC2D}), these time limits were longer than the time to completion
all algorithms in simulation. Additional motivation for limiting the run
time were constraints on the physical experiment, and the observation
that when algorithms failed they tended to fail in such a way that the
estimate variance never fell below a certain threshold (excluding the
random walk controller), and the success criteria listed above could not
be applied.

\section{Trial scenarios \& results}
\label{results}

Experiments were designed to determine whether the EEDI algorithm
performs at least as well as several alternative choices of controllers
in estimating of the location of stationary target(s), and whether there
were scenarios where EEDI outperforms these alternative controllers,
e.g. in the presence of distractor objects or as the number of targets
increases. Experiments in Sections \ref{POC1D} through
\ref{2dtargetradius} are performed using the kinematically controlled
SensorPod robot and simulation, and these results are summarized in
Section \ref{expsummary}. In Section \ref{dynamics}, we transition to
simulation-only trials to demonstrate successful closed-loop estimation
of target location, but compare trajectories and performance using three
models of the robot; the kinematic model of the experimental system, a
kinematic unicycle model, and a dynamic unicycle model.

\label{altalgs}
In Sections \ref{POC1D} through \ref{2dtargetradius} we compare the performance
of  EEDI  to the following three  implementations of information maximizing controllers and a random walk controller:
\begin{enumerate}[I.]
\item \textbf{Information Gradient Ascent Controller (IGA)} The IGA
  controller drives the robot in the direction of the gradient of the
  EID at a fixed velocity of 4 cm/s, inspired by
  controllers used in \cite{Grocholsky06, kreucher2007, lu11,
    Bourgault02i}.
\item \textbf{Information Maximization Controller (IM)} The SensorPod is
  controlled to the location of the EID maximum, at a constant velocity
  for time $T$, similar to \cite{Li05,liao04,Wong05,VanderHook2012}.
\item \textbf{Greedy Expected Entropy Reduction (gEER)} At each
  iteration, fifty locations are randomly selected, within a fixed
  distance of the current position. The SensorPod is controlled to the
  location that maximizes the expected change in entropy, integrated
  over the time horizon T.\footnote{The expected entropy reduction is
    $H(\theta)-E[H(\theta) | V^+(t)]$ where
    $H(\theta)=-\int p(\theta) \log p(\theta)d\theta$ is the entropy of
    the unknown parameter $\theta$ \cite{thrun05p,Tisd09a} and $V^+(t)$
    is the expected measurement, calculated for each candidate
    trajectory $x^+(t)$, the current estimate $p(\theta)$, and the
    measurement model.} This approach is similar to the method of
  choosing control actions in \cite{Fox98, kreucher05s,
    Feder99,souza2014}
\item \textbf{Random Walk (RW)} The SensorPod executes a randomly
  chosen, constant velocity trajectory from the current sensor position
  for time $T$, similar to \cite{Solb08a}.
\end{enumerate}
The planning horizon $T$ was the same for all controllers, so that the
same number of measurements is collected.

Alternative algorithms, for example a greedy gradient-based controller
(IGA) or a random walk (RW), produce control signals with less
computational overhead than the EEDI algorithm because the
EEDI involves solving a continuous trajectory optimization problem and
evaluating an additional measure on ergodicity. In the next section we
demonstrate several scenarios in which the tradeoff in computational
cost is justified if the estimation is likely to fail or suffer
significantly in terms of performance using less expensive control
algorithms. Additionally, the alternative algorithms, while appropriate
for the kinematically-controlled SensorPod robot, do not generalize in
an obvious way to nonlinear dynamic models. This is one of our reasons
for desiring a measure of nonmyopic search that can be expressed as an
objective function (i.e. ergodicity). Given an objective, optimal
control is a convenient means by which one makes dynamically dissimilar
systems behave similarly to each other according to a metric of
behavior. In the case of exploration, the measure is of coverage
relative to the EID---however it is constructed.

\subsection{Performance comparison for 1D target estimation in the
  presence of an unmodeled distractor}\label{POC1D}
In this section, the robot motion is constrained to a single dimension,
and the estimation objective is the 1D location $\theta$ of a single,
stationary target with known radius. A {\it distractor object} is placed
in the tank, as an unmodeled disturbance, in addition to the (modeled)
{\it target object}. Both the target and the distractor were identical
non-conductive, 2.5 cm diameter spheres, placed at different fixed
distances from the SensorPod's line of motion (see Fig. \ref{1dtanks}).
The voltage signal from the distractor object is similar but not
identical to that of the target (see Fig. \ref{fig:voltagetrace}).
Placing the distractor object further from the SensorPod line of motion
results in decreased magnitude and rate of change of the voltage trace.
Introducing an unmodeled distractor even in a one-dimensional sensing
task was enough to illustrate differences in the performance of the EEDI
Algorithm and Algorithms I-IV.
 
  \begin{figure}[t]
   \centering  
 \vspace{-.0cm}
 \subfloat[Simulation ]
 {\label{FIG10_2}\includegraphics[trim=.58in
    .1in .5in
    .09in,clip=true,width=0.5\columnwidth ]{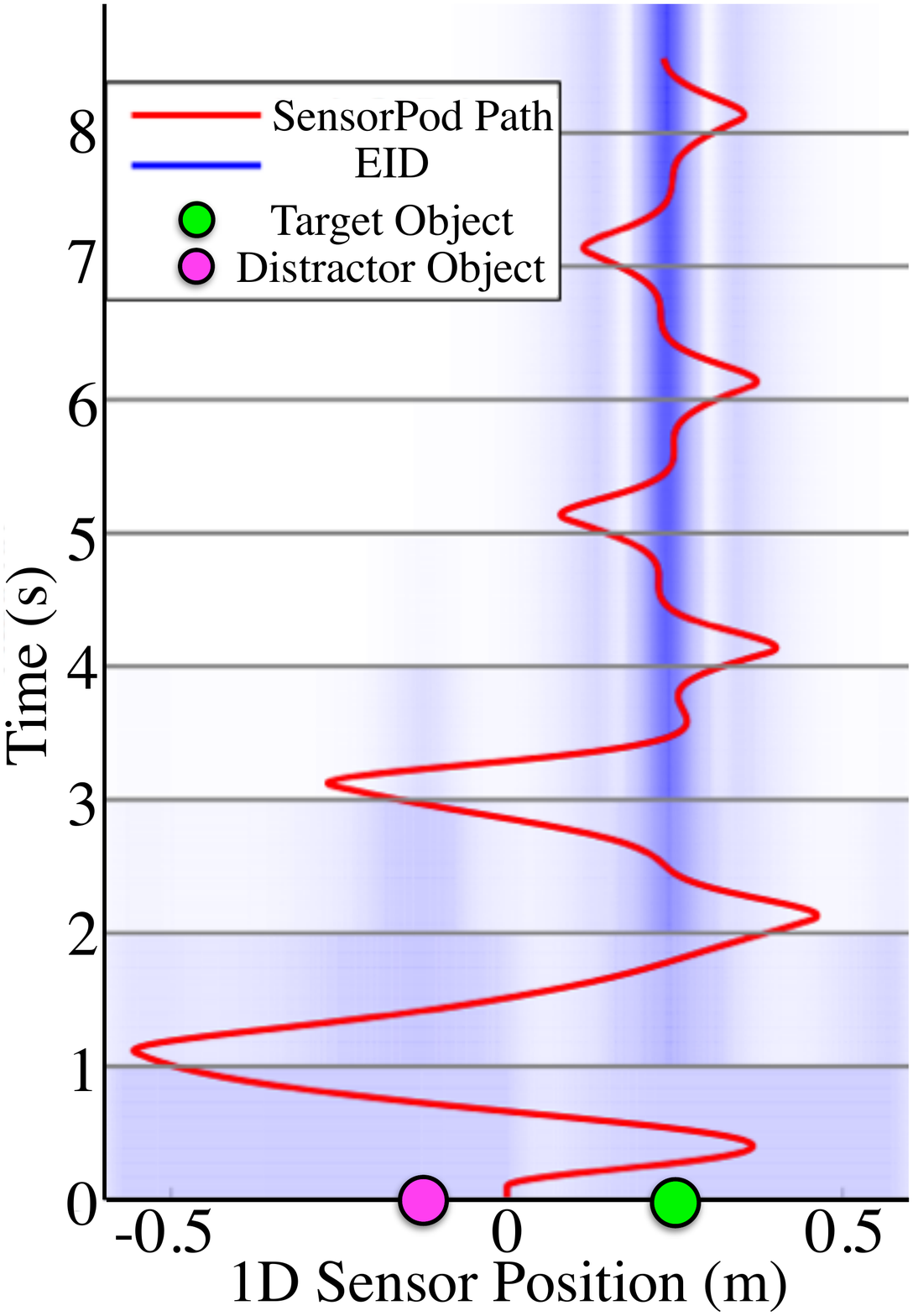}}
 \subfloat[ Experiment ] {\label{FIG10_1}\includegraphics[trim=.32in
    .1in .3in
    .09in,clip=true,width=0.5\columnwidth ]{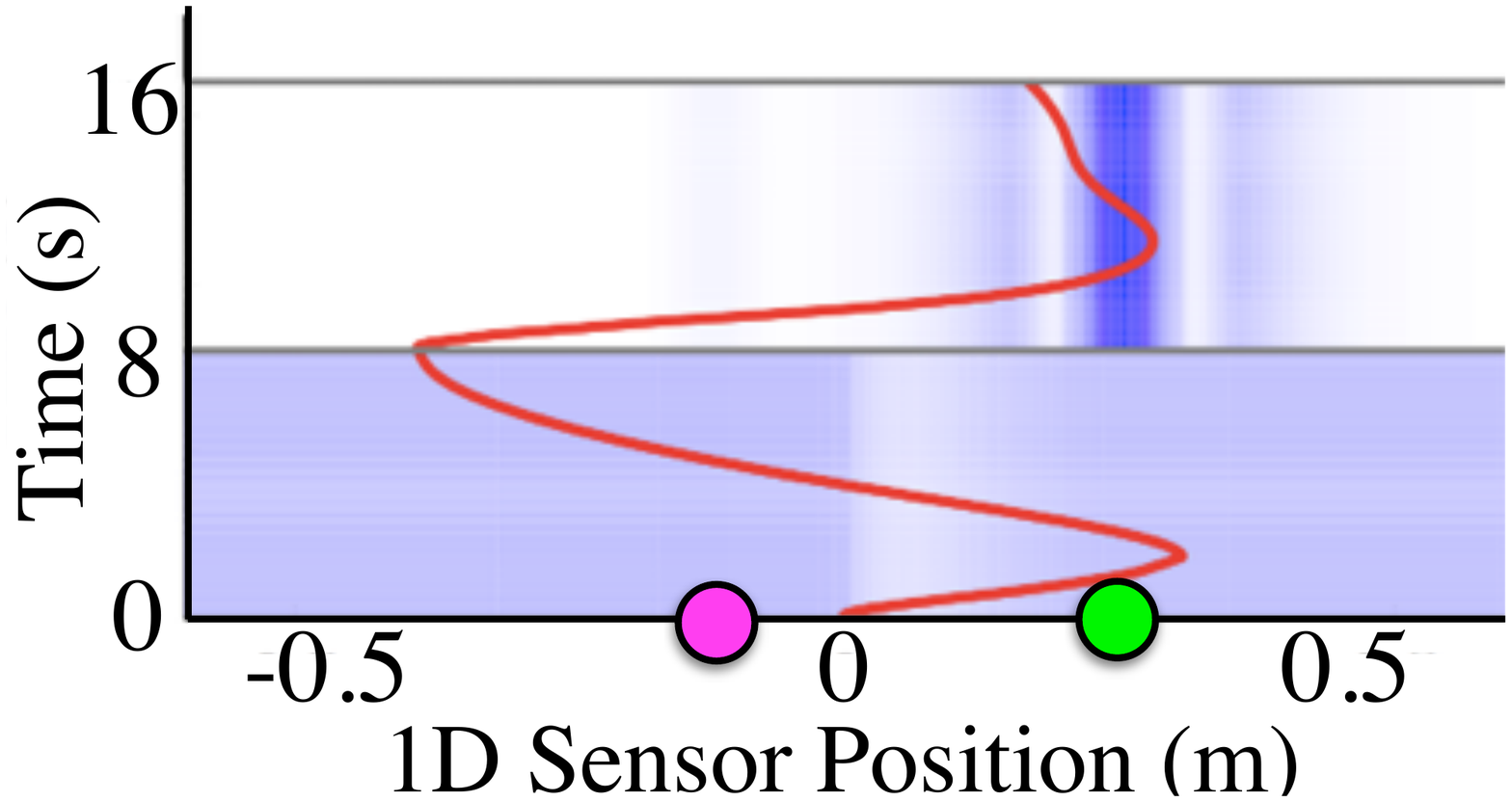}}
  \caption{Examples of closed-loop optimally ergodic search in
    simulation and experiment. The EID is shown as a density plot in
    blue, and the search trajectory in red. The belief and trajectory are
    recalculated every second in simulation and every 8 seconds in
    experiment.\vspace{-10 pt}}
 \label{1dtraj1} 
 \vspace{-0cm}
 \end{figure}

\begin{table}[t]
  \caption{
    Performance measures for estimation of single target location in 1D for 100 simulated and 10 experimental trials.  Results for time until completion (slowdown factor) are only shown for successful trials. Slowdown factor of 1 corresponds to 15.2 s in experiment, 7.6 s in simulation.}
\begin{center}
    \begin{tabular}{|c|c | c| c|c|c|} \hline Description & EEDI&gEER &IM&IGA&RW \\
      \hline \hline
      Exp. Success \%  & 100 &50  & 60 & 50 & 80 \\
      Sim. Success  \% &100  & 60 & 71 & 66 & 99 \\\hline 
      Exp. Slowdown Factor    & 1  &1.4  &2.1  & 2.7 &2.7 \\ 
      Sim. Slowdown Factor    & 1   &2.1  &2.1  & 2.3 & 6.3\\ \hline
\end{tabular}
\end{center}
\label{tableBIG} \vspace{-15 pt}
 \end{table}

 \begin{figure*}[t]\vspace{-0 pt}
 \centering 
 \includegraphics[
    width=1\textwidth ]{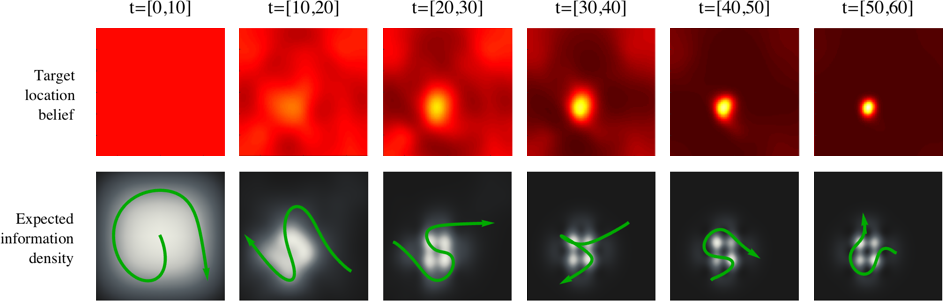}\vspace{-15pt}
 \caption{A progression of the estimate of the two-dimensional target
   location using the EEDI algorithm. As the algorithm progresses,
   collected measurements evolve the estimate from a uniform
   distribution over the workspace (top-leftmost figure), to a
   concentrated distribution at the correct location. At each interval,
   the EID is calculated from the updated estimate, which is then used
   to calculate an ergodic search trajectory. \vspace{-10
     pt}}\label{2dtraj}
 \end{figure*}

 We performed 100 trials in simulation and 10 in experiment, with the
 target position held constant and the distractor's position along the
 SensorPod's line of motion randomized.\footnote{ The only additional
   consideration in the experimental scenario was that the tank walls
   and the water surface have a non-negligible effect on measurements.
   We compensate for this by collecting measurements on a fine grid in
   an empty tank, and subtracting these measurements at each measurement
   point during estimation.} The results for success rate and average
 slowdown factor (for successful trials), averaged over all trials in
 simulation and experiment, are summarized in Table \ref{tableBIG}. The
 slowdown factor is the average time until completion, normalized by the
 minimum average time until completion in experiment or simulation.
 Results presented in Table \ref{tableBIG} demonstrate that the EEDI
 algorithm localizes the target successfully 100\% of the time, and does
 so more quickly than Algorithms I-IV.

 Differences in time to completion between experimental and simulated
 trials are due to experimental velocity constraints. In simulation, the
 time horizon used for trajectory optimization, and therefore between
 PDF updates, was one $T=1$ second. A longer ($T=8$ seconds) time
 horizon was used for experimental trajectory optimization, avoiding the
 need to incorporate fluid dynamics in the measurement model; at higher
 velocities the water surface is sufficiently disturbed to cause
 variations in the volume of fluid the field is propagating through,
 causing unmodeled variability in sensor measurements.

Figure \ref{1dtraj1} shows experimental and simulated examples of
closed-loop one-dimensional trajectories generated using the EEDI
algorithm. Given no prior information (a uniform belief), the ergodic
trajectory optimization initially produces uniform-sweep-like behavior.
In the experimental trial shown in Fig. \ref{FIG10_1}, the termination
criteria on the variation of the PDF is reached in only two iterations
of the EID algorithm, a result of the longer time horizon and resulting
higher density measurements. The distributed sampling nature of the EEDI
algorithm can be better observed in the simulated example shown in Fig.
\ref{FIG10_2}, where shorter time horizons and therefore more sparse
sampling over the initial sweep require more iterations of shorter
trajectories. As the EID evolves in Fig. \ref{FIG10_2}, the shape of the
sensor trajectory changes to reflect the distribution of information.
For example, the sensor visits the local information maximum resulting
from voltage perturbations due to the target and the local information
maximum due to the distractor between 1 and 4 seconds. Experimental
results for this trial were presented in \cite{silverman13}.

\begin{table}[t]\vspace{-0 pt}
\caption{Performance measures for estimation of
single target location in 2D for 10 simulated and 10 experimental
trials. Results for time until completion (slowdown factor) are only shown for successful trials. Slowdown factor of 1 corresponds to 64 s in experiment, 65.2 s in simulation.}
\begin{center}
\begin{tabular}{|c|c | c| c|c|c|}
  \hline
  Description & EEDI &gEER&RW \\
 \hline
\hline
 Exp. Success \%  & 100 & 90 & 90 \\
Sim. Success \%   &100 & 100 & 100 \\
  \hline
Exp. Slowdown Factor& 1 &1.2 & 2.9\\
  Sim.  Slowdown Factor & 1 &1.1 & 2.6\\
\hline
\end{tabular}
\end{center}\vspace{-15 pt}
\label{table2DND}
\end{table}

 \subsection{Performance comparison for estimating the 2D location of
   a single target} \label{POC2D}

 In this section, the robot is allowed to move through a
   2D workspace and the objective was to compare the performance of EEDI
   to gEER and RW for 2D, stationary target localization, i.e.
   $\bm \theta=(\theta_x,\theta_y)$. No distractor object was present as
   the difference in performance between algorithms was notable even
   without introducing a distractor object. Fig. \ref{2dtanks} shows an
   example tank configuration for multiple target estimation in 2D. We
   omit comparison to IGA and IM for 2D experiments; RW is the simplest
   controller to calculate and resulted in high success percentage for
   1D trials, and gEER, with performance similar to IGA and IM on
   average in 1D trials, is qualitatively similar to our approach and
   more commonly used.

Ten trials were run for each of the EEDI, gEER, and RW algorithms, both
in experiment and simulation, with the target location randomly chosen.
Figure \ref{2dtraj} shows the convergence of the belief at 10 second
intervals ($T=10$), as well as the corresponding EID and ergodic
trajectory. The performance measures for experimental and simulated
trials using the EEDI, gEER, and RW algorithms are shown in Table
\ref{table2DND}. In simulation, all three algorithms have 100\% success
rate, while the gEER and RW controllers have a 10\% lower success rate
in the experimental scenario. The gEER controller requires roughly
10-20\% more time to meet the estimation criteria, whereas the RW
controller requires about 2-3 times more time. As mentioned in the
previous section, although gEER performs well in this scenario, it did
not perform as well with distractors.

\subsection{Performance comparison for estimating the 2D location of
  multiple targets} \label{POCmultiple} 
\begin{figure}[!t]\vspace{-15 pt}
   \centering 
\subfloat[Experiment] {\label{multplotsa}\includegraphics[trim=.05in
    .051in .05in
    .1in,clip=true,width=0.5\columnwidth ]{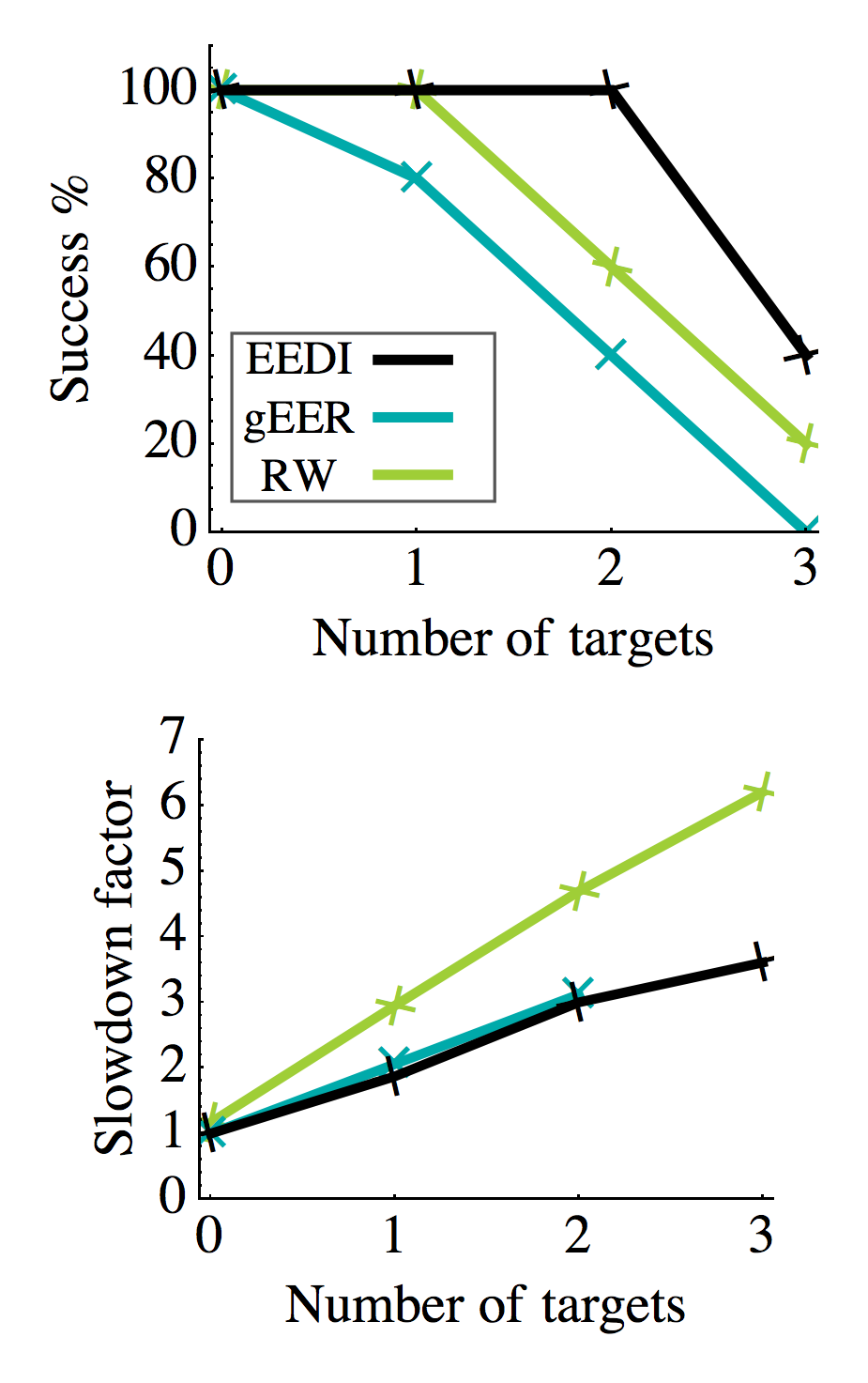}}
 \subfloat[Simulation] {\label{multplotsb}\includegraphics[trim=.05in
    .051in .05in
    .1in,clip=true,width=0.5\columnwidth ]{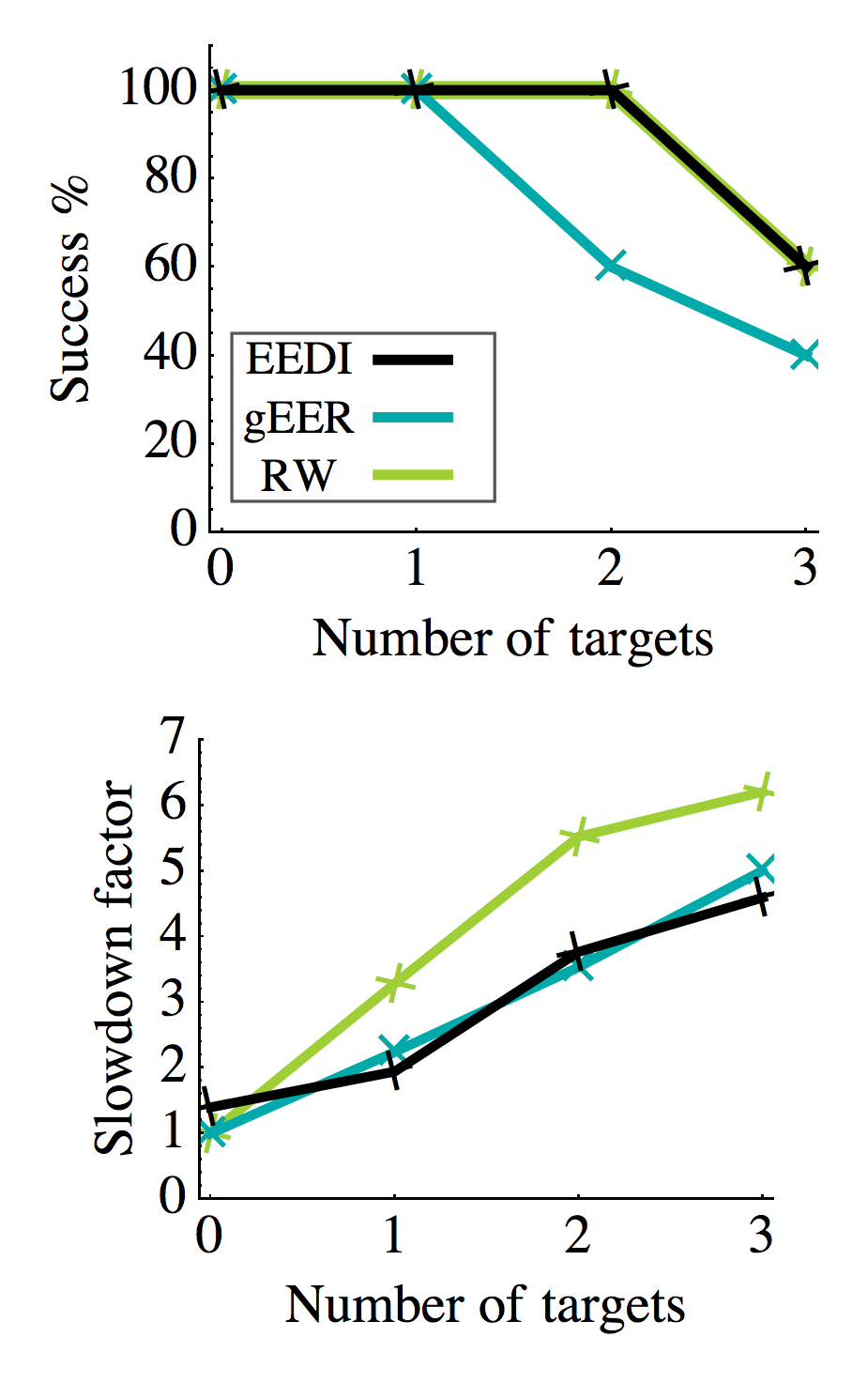}}
 \caption{Performance measures for estimation of multiple target
   locations in 2D for five experimental and
   five simulated trials. Slowdown factor of 1 corresponds to 50 seconds in simulation, 60 seconds in experiment.\vspace{-00 pt}}
 \label{multplots}
 \vspace{-0cm}  
 \end{figure}

Having demonstrated that the
EEDI algorithm modestly outperforms gEER and drastically outperforms RW
(in terms of time) for localizing a single, stationary target in 2D, we
next sought to compare EEDI performance localizing multiple targets (see
Fig. \ref{2dtanks}). We compare the EEDI algorithm to the gEER and RW
controllers, again leaving out IM and IGA because of their poor
performance in Section \ref{POC1D}. We performed localization estimation
for scenarios where there were either 0, 1, 2, or 3 targets in the tank,
all 2.5 cm diameter. We conducted 5 trials in simulation and experiment,
for each number of targets (with all locations selected randomly).
Figure \ref{multplots} shows the percentage of successful trials and
average slowdown factor as a function of the number of targets in the
tank. The slowdown factor is calculated by normalizing average time
until completion by the minimum average time until completion for all
algorithms and all target numbers.

In the experimental scenario, Fig. \ref{multplotsa}, the EEDI
algorithm had a higher success rate than both the gEER and RW
controllers for higher numbers of objects. The slowdown factor
using the EEDI algorithm was very similar to the gEER algorithm for
0-2 objects (the gEER controller never successfully localized 3
objects), and much shorter than the RW controller. In simulation, Fig.
\ref{multplotsb}, the success rate of the EEDI algorithm matched that
of the RW, however the RW slowdown factor was much greater.

\begin{figure}[t]
  \begin{center}
    \vspace{-0 in}
    \includegraphics[width=.7\columnwidth ]{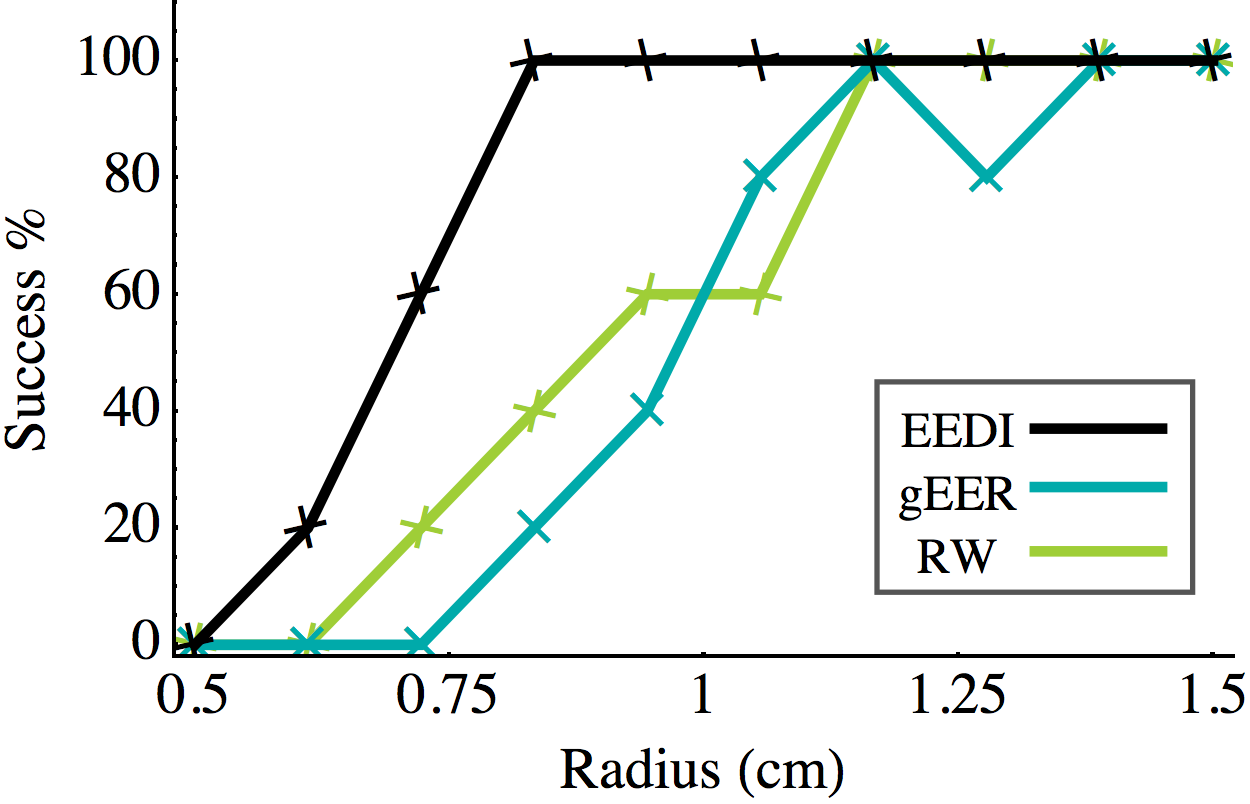}
  \end{center}
  \caption{Success Rate for estimation of location and radius, as a function of target radius, for simulated trials only.  \vspace{-10 pt} }\label{figss}
\end{figure} 
\subsection{Performance degradation  with signal to noise ratio}\label{2dtargetradius}

The next trial is used to demonstrate an extension of the EEDI Algorithm
to non-location parameters, and to examine performance degradation as a
function of the signal to noise ratio. As mentioned, the EEDI algorithm
can also be used to determine additional parameters characterizing
target shape, material properties, etc. The only requirement is that
there be a measurement model dependent on---and differentiable with
respect to---that parameter. Parameters are incorporated into the
algorithm in the same way as the parameters for the spatial location of
a target (see Appendix \ref{Fisher Information}). We therefore
demonstrate effectiveness of the EEDI algorithm for estimation of target
radius as well as 2D target location. We estimated target location and
radius for ten different radii varying between 0.5 cm to 1.5 cm. Five
trials were performed for each target radius, with randomly chosen
target locations. By varying the radius of the target, which for our
sensor results in scaling the signal to noise ratio,\footnote{the signal
  drops approximately with the fourth power of the distance from a
  spherical target, and increases with the third power of target radius
  \cite{Nelson2006}} we are able to observe relative performance of
several comparison algorithms as the signal to noise ratio drops off.
Trials were performed in simulation only. Figure \ref{figss} shows the
mean success rate of the five simulated trials as a function of target
radius. For EEDI, gEER, and RW the success rate decreased as the radius
decreased. This is expected, as the magnitude of the voltage
perturbation, and therefore the signal to noise ratio, scales with
$r^3$. For objects with $r<0.9$ cm, the peak of the expected voltage
perturbation is less than the standard deviation of the sensor noise.
Nevertheless, the EEDI Algorithm had a higher success rate than gEER and
RW for radii between 0.5 cm and 1 cm.

\begin{figure}[t] 
   \centering 
 \vspace{-0pt}
 \subfloat[\vspace{-10 pt}] {\label{edittext1Da}\includegraphics[width=0.7\columnwidth ]{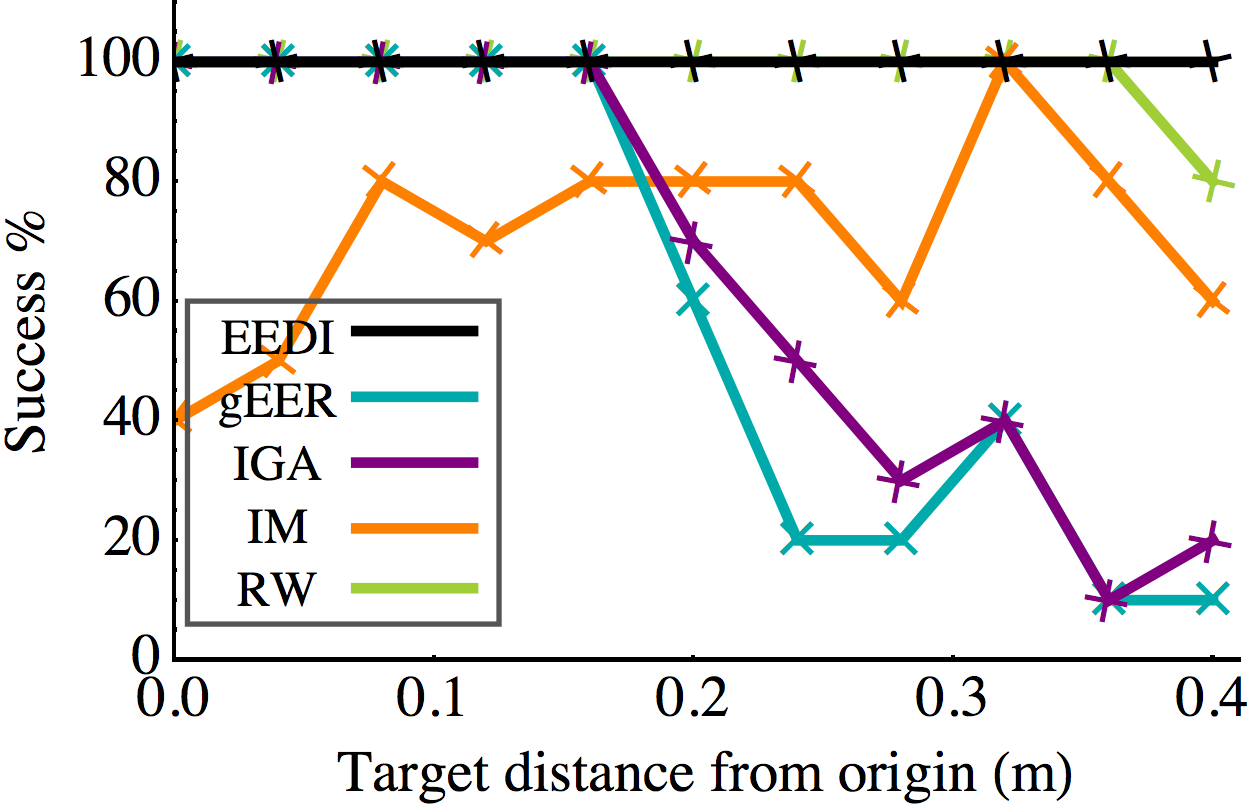}}\\
 \subfloat[] {\label{edittext1Db}\includegraphics[width=0.7\columnwidth ]{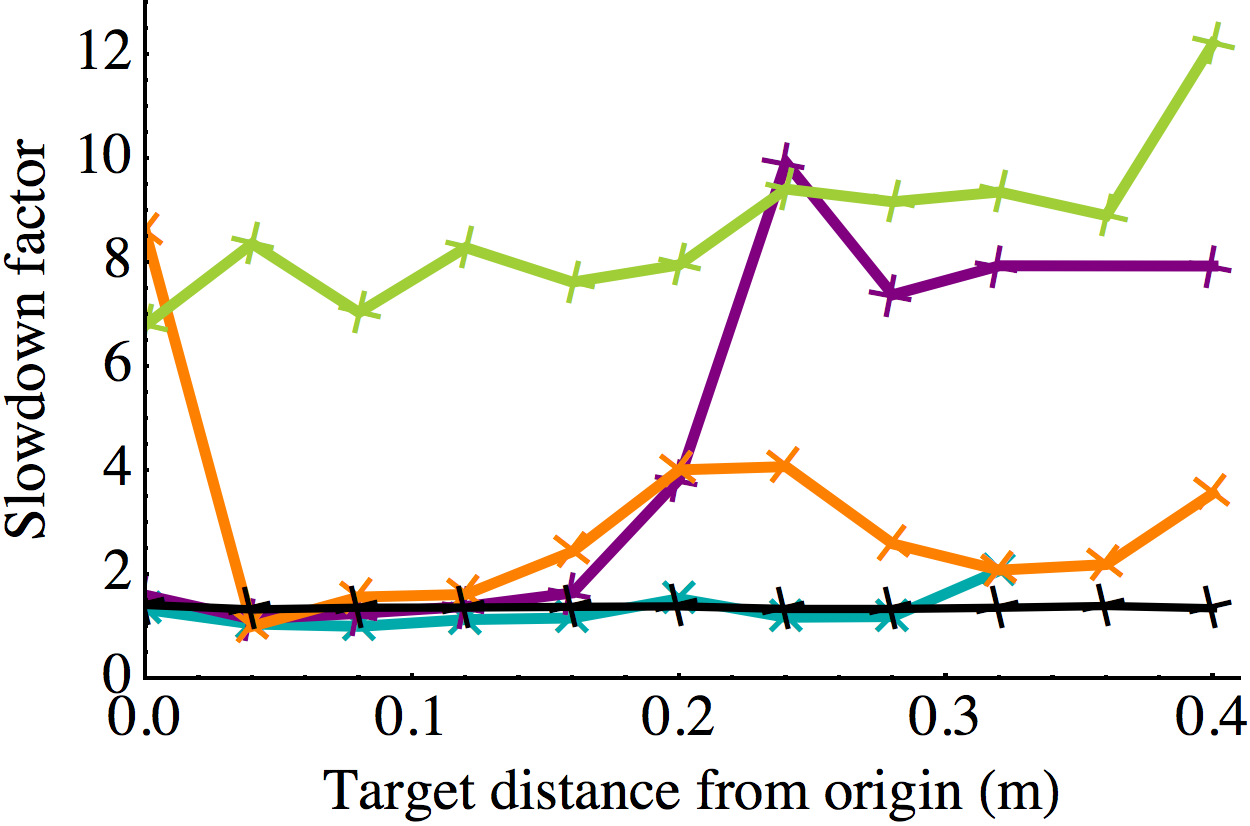}}
 \caption{Performance measures for estimation of single target location
   in 1D are shown for the EEDI algorithm and Algorithms I-IV. Results
   of 110 simulated trials are shown for each algorithm; for each of 11
   target locations, 10 simulated trials were performed with the 1D
   distractor object location randomized (a fixed distance from the
   SensorPod line of motion was maintained). A slowdown factor of 1
   corresponds to 5.61 seconds; slowdown factor is not shown for target
   distances with less than 10\% success rate.\vspace{-10 pt}}
 \label{edittext1D}
 \vspace{-0cm}
 \end{figure}

\subsection{Comparison of sensitivity to 
  initial conditions }\label{IC}

Finally, we use the one-dimensional estimation scenario (the same as
that in Section \ref{POC1D}) to illustrate the relative sensitivity of
the EEDI algorithm and Algorithms I-IV to the initial conditions of the
sensor with respect to locations of the target and an unmodeled
disturbance. This captures the likelihood of different controllers to
become stuck in local minima resulting from the presence of a distractor
object which produces a measurement similar but not identical to the
target.

We executed a total of 110 simulated trials for each algorithm. 10
trials were simulated for 11 equally spaced target locations. For each
target location, the distractor location was randomized, with a minimum
distance of 25 centimeters distance from the target (along the SensorPod
line of motion, to prevent electrosensory occlusion). 110 trials allowed
significant separation of the results from different controllers. For
all trials, the SensorPod position was initialized to $(x,y) = 0$.

Figure \ref{edittext1D} shows the performance measures for Algorithms
I-IV. The slowdown factor is calculated by normalizing average time
until completion by the minimum average time over all algorithms and all
target locations. When the target was located near the SensorPod initial
position, EEDI, gEER, IGA, and RW perform comparably in terms of success
percentage and time, with the exception being the RW controller, which
is predictably slower. Success rate drops off using gEER and IGA for
target positions further from the SensorPod initial position. Note that
IM performs poorly if the target is located exactly at the robot start
position, due to the nonlinear characteristics of electrolocation. A 0 V
measurement would be observed for a target located at the sensor
position or anywhere sufficiently far from the sensor; this means that
the initial step of the IM algorithm has a high probability of driving
the sensor to a position far from the target. EEDI, on the other hand,
localized the target in essentially constant time and with 0\% failure
rate regardless of location. The RW algorithm performs as well as the
EEDI algorithm in terms of success rate, but is approximately eight
times slower.

\subsection{Summary of experimental results}\label{expsummary}
In Sections \ref{POC1D} and \ref{POC2D}, we provide examples of
successful estimation trajectories for the EEDI algorithm. In the
two-dimensional estimation problem in Section \ref{POC2D}, we observe
that both success rate and time until completion are comparable using
both EEDI and gEER algorithms (with time being much longer for the
random walk controller). While this scenario illustrates that our
algorithm performs at least as well as a greedy algorithm in a simple
setting, and more efficiently than a random controller, where we really
see the benefit in using the EEDI algorithm is when the robot is faced
with more difficult estimation scenarios. Experiments in Section
\ref{IC} showed that the EEDI algorithm was robust with respect to
initial conditions (i.e. whether or not the sensor happens to start out
closer to the distractor or target object) where Algorithms I-IV are
sensitive. For Algorithms I-IV, the further the target was from the
initial SensorPod position, the more likely the estimation was to fail
or converge slowly due to local information maxima caused by the
distractor. Similarly, when the estimation objective was target
localization for varying numbers of targets in Section \ref{POCmultiple}
(a scenario where many local information maxima are  expected), the
success rate of the EEDI algorithm is higher than expected entropy
reduction and completion time is shorter than the random walk as the
number of targets increased. Lastly, the success rate of the EEDI
algorithm degraded the least quickly as the signal to noise ratio
declined. In addition to outperforming alternative algorithms in the
scenarios described, the ergodic trajectory optimization framework
enables calculation of search trajectories for nonlinear,
dynamically-constrained systems.

\subsection{Comparison of different dynamic models}\label{dynamics}
One of the benefits of ergodic optimal control is that the control
design does not change when we switch from a kinematic robot model to a
dynamic robot model. While the physical SensorPod robot is controlled
kinematically due to the gantry system, we can simulate nonlinear and
dynamic models to see how dynamics might influence information gathering
during untethered movement for future SensorPod iterations. We simulate
automated target localization using the EEDI algorithm for the SensorPod
robot using three different models for the robot dynamics. All
parameters in the ergodic optimal control algorithm are exactly the same
in all three cases: the weights on minimizing control effort vs.
maximizing ergodicity, in Eq. \eqref{Jdis2}, were set to $\gamma = 20$,
$R =0.01 \mathbb I$, (where $\mathbb I$ is a $2 \times 2$ identity
matrix), and the planning time horizon was $T=10$. In all three cases
below, the measurement model was identical and defined relative to the
X-Y position of the robot, although the system state differs. The only
changes in the implementation are the robot's state and equations of
motion for the three systems, defined below.

\begin{figure}[!t] 
   \centering 
 \vspace{-0pt}
 \subfloat[Linear, kinematic system \vspace{-0 pt}] {\label{simkinematic}\includegraphics[trim=.15in .0in .2in
     .0in,clip=true,width=\columnwidth ]{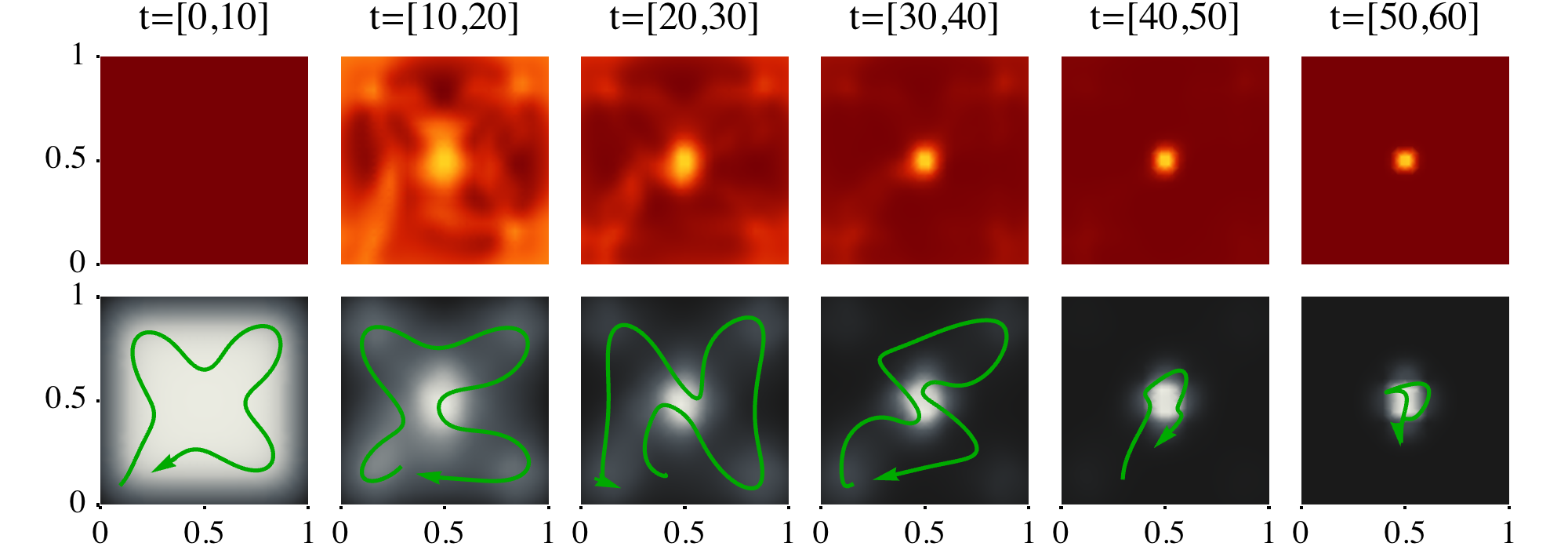}}\\
 \subfloat[Kinematic unicycle model (nonlinear, kinematic system)] {\label{simnonlinear}\includegraphics[trim=.15in .0in .2in
     .0in,clip=true,width=\columnwidth ]{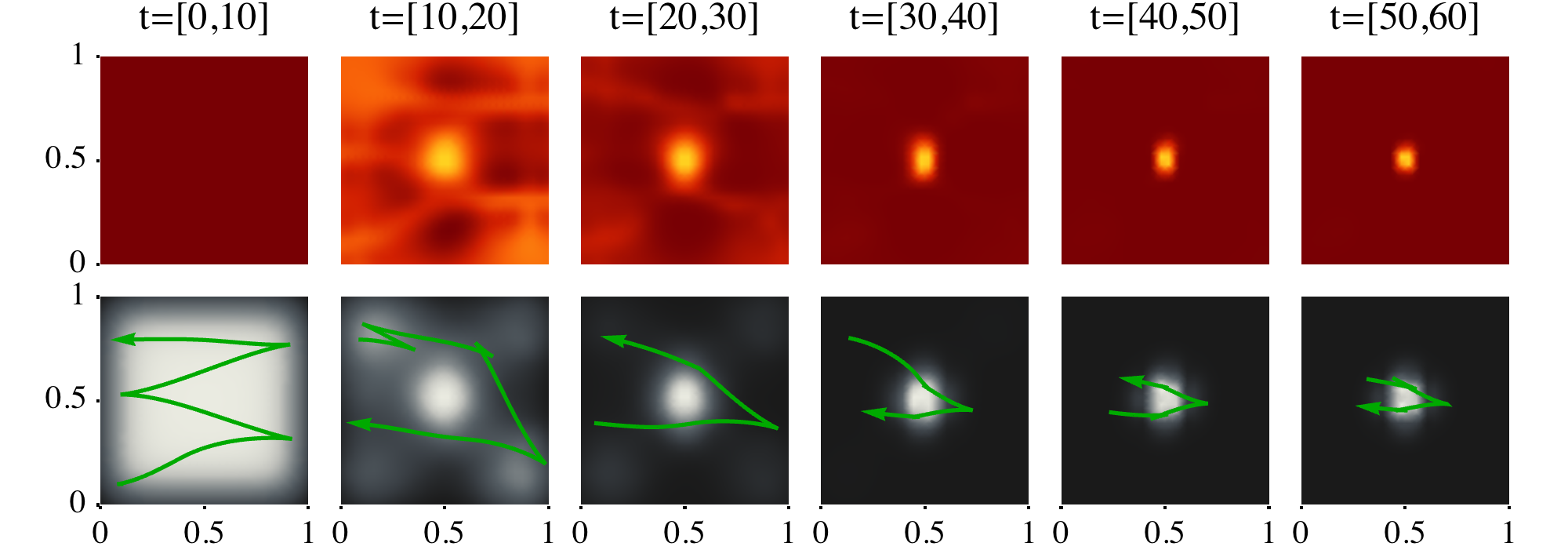}}\\
\subfloat[Dynamic unicycle model (nonlinear, dynamic system)\vspace{-
0 pt}] {\label{simdynamic}\includegraphics[trim=.15in .0in .2in
     .0in,clip=true,width=\columnwidth ]{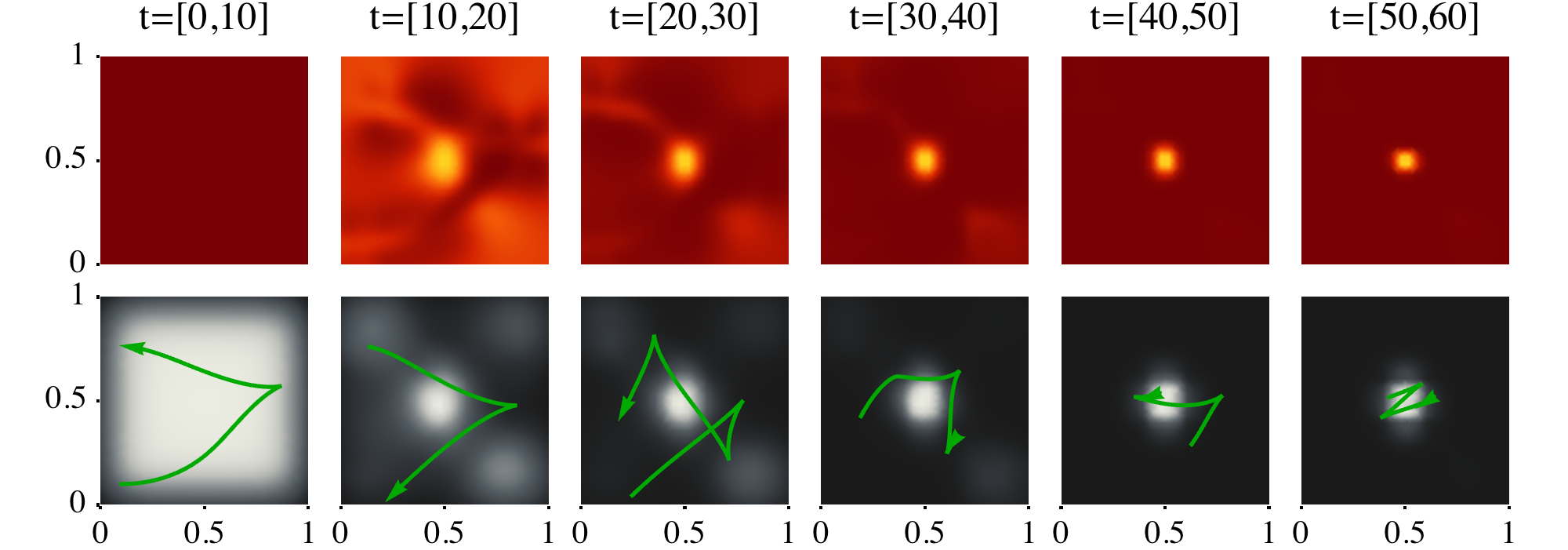}}
   \caption{A progression of the estimate of the two-dimensional target
     location using the EEDI algorithm, in simulation, for three
     different systems performing the same task. As the algorithm
     progresses, collected measurements evolve the estimate (heatmap)
     from a uniform distribution over the workspace (top-leftmost figure
     in each (a),(b),(c)), to a concentrated distribution at the correct
     location. At each interval, the EID (grayscale) is calculated from
     the updated estimate, which is then used to calculate an ergodic
     search trajectory (green).\vspace{-0 pt}}
 \label{simdiffsystems}
 \vspace{-0cm}
 \end{figure}

\begin{figure}[!t]
   \centering 
 \vspace{-0pt}
 \includegraphics[width=.7\columnwidth ]{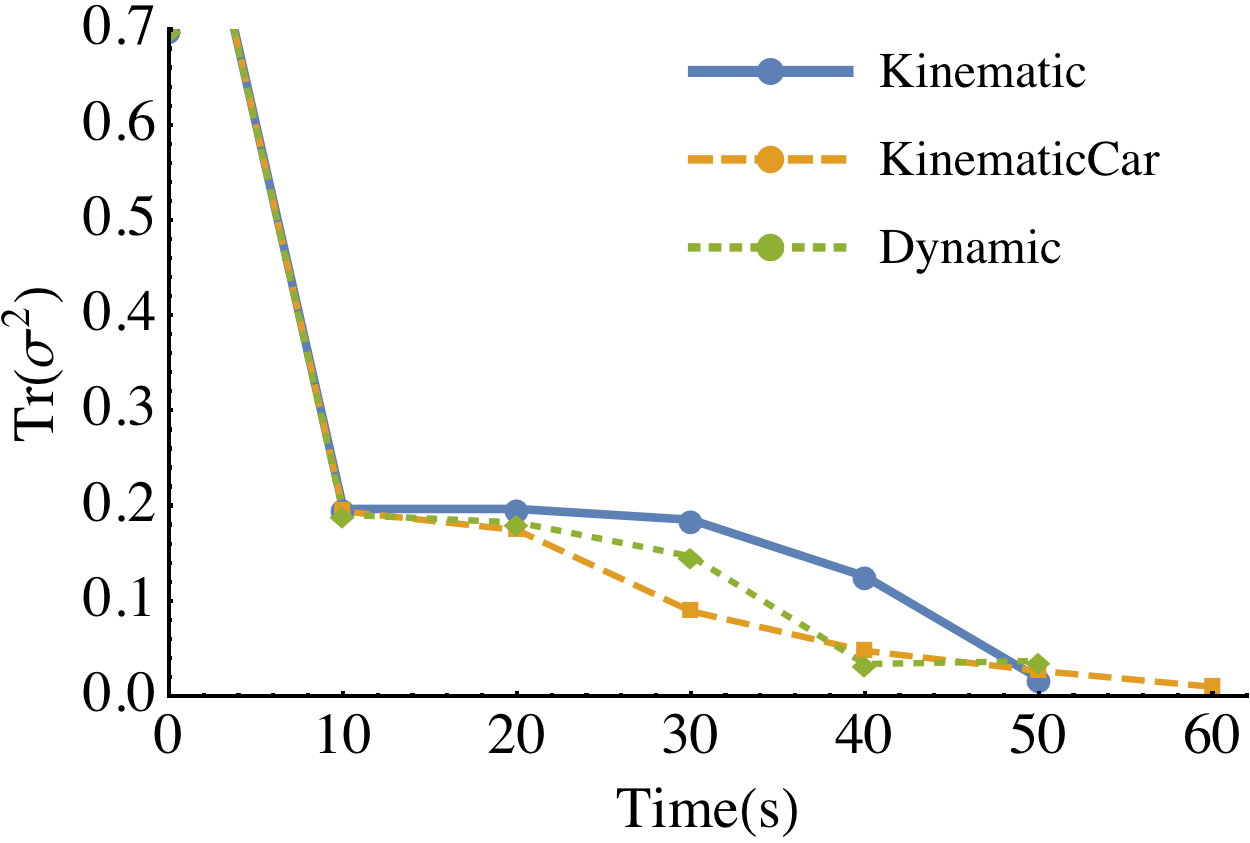}
 \caption{The trace of the covariance of the two-dimensional target
   location estimate is plotted as a function of time. We observe
   similar overall convergence behavior for all three systems for this
   particular set of initial conditions and weighted objective function.
   The covariance is updated after executing each 10-second long
   trajectory. \vspace{-10 pt}}
 \label{systemconvergenceplot}
 \vspace{-0cm}
 \end{figure}

\subsubsection{Linear kinematic system}
The state is $\bm x(t) = (x(t),y(t))$ where $ x(t)$ and $y(t)$ are Cartesian
coordinates, and the equations of motion are  $\bm{\dot{x}(t)}=\bm u(t).$ The initial conditions were $\bm x(0) = (0.1, 0.1)$.

\subsubsection{Nonlinear kinematic system}
We use the standard kinematic unicycle model. The state is
$ \bm x(t)=(x(t),y(t),\theta(t))$ where $x(t)$ and $y(t)$ are Cartesian
coordinates and $\theta(t)$ is a heading angle, measured from the $x$
axis in the global frame. The control $ \bm u(t)=(v(t), \omega(t))$
consists of a forward velocity $v(t)$ and angular velocity $\omega(t)$.
The equations of motion are \begin{align}\label{dyneq}
  \dot{\bm{x}}(t)=\begin{bmatrix}
    \dot x(t)  \\
    \dot y(t)  \\
    \dot \theta(t)
 \end{bmatrix}=\begin{bmatrix} v(t) \cos\theta(t) \\ v(t)\sin\theta(t) \\ \omega(t)
 \end{bmatrix}.
 \end{align}
The initial conditions were $\bm x(0) = (0.1, 0.1, 0)$.

\subsubsection{Nonlinear Dynamic System}
We use a dynamic variation on the unicycle model. In this case the state
is $\bm x(t)=(x(t),y(t),\theta(t),v(t),\omega(t))$ where
$x,y,\theta,v,\omega$ are the same as in the kinematic unicycle model.
The control inputs are $ \bm u(t) = (a(t),\alpha(t))$, with the
equations of motion \begin{align}\label{dyneq}
  \dot{\boldsymbol{x}}(t)=\begin{bmatrix}
    \dot x(t)  \\
    \dot y(t)  \\
    \dot \theta(t)\\
    \dot v(t)\\
    \dot \omega(t)
 \end{bmatrix}=\begin{bmatrix} 
v(t) \cos\theta(t) \\ 
v(t) \sin\theta(t) \\
\omega(t)\\
\tfrac{1}{2} a(t)\\
 \alpha(t)
 \end{bmatrix}.
       \end{align}
       The initial conditions were $\bm x(0) = (0.1, 0.1, 0, 0, 0)$.
       Figure \ref{simdiffsystems} illustrates the progression of the
       EEDI algorithm for static, single target localization for all
       three systems. In all cases, we use a finite time horizon of
       $T=10$ seconds for trajectory optimization, and the PDF is
       initialized to a uniform distribution. While the types of
       trajectories produced are qualitatively different because of the
       different dynamic constraints, we observe similar convergence
       behavior for all three systems for this particular set of initial
       conditions and weights in the objective function. In Fig.
       \ref{systemconvergenceplot}, the trace of the estimate covariance
       is plotted as a function of EEDI iterations.

\section{Conclusion}
\label{conclusion}
We present a receding horizon control algorithm for active estimation
using mobile sensors. The measurement model and belief on the estimates
are used to create a spatial map of expected information gain. We
implement our algorithm on a robot that uses a bio-inspired sensing
approach called electrolocation \cite{Neve13a}. Ergodic trajectory
optimization with respect to the expected information distribution, as
opposed to information maximization, is shown to outperform alternative
information maximization, entropy minimization, and random walk
controllers in scenarios when the signal to noise ratio is low or in the
presence of disturbances.

One major advantage of ergodic trajectory optimization is that the
formulation is suitable for systems with linear or nonlinear, kinematic
or dynamic motion constraints, as shown in Section \ref{dynamics}.
Additionally, the method does not formally rely on discretization of the
search space, the action space, or the belief space. Although numerical
integration schemes are used in solving differential equations or
updating the belief, discretization is an implementation decision as
opposed to a part of the problem statement or its solution. Another
benefit is that neither assuming information submodularity \cite{Sim05,
  Singh2009,Hollinger2014}) nor selecting waypoints \cite{zhang09B,
  zhang09} are required to distribute measurements among different
regions of high expected information when planning over a long time
horizon. Using ergodicity as an objective also means that the algorithm
is suitable for both coverage \cite{Acar2003, choset2001} or ``hotspot''
sampling, without modification. If the information density is very
concentrated, the optimally ergodic trajectory will be similar to an
information maximizing solution.\footnote{Note that this would only
  happen for measurement models that cause the EID to converge to a
  low-variance, unimodal distribution that approximates a delta function
  (where the equivalence between an information maximizing solution and
  an ergodic solution follows directly from their definitions); this
  does not happen in the examples shown in Section \ref{results}.
  Because of the highly nonlinear measurement model, the EID converges
  to a multimodal density function, as shown in Fig \ref{FItight}.} On
the other hand, if the information density is diffuse (or the planning
time horizon very long), the optimally ergodic solution will approximate
a coverage solution. In Figs. \ref{2dtraj} and \ref{simdiffsystems},
coverage-like solutions are observed for the initial, nearly-uniform
belief; although the belief converges, the EID does not converge to a
unimodal distribution due to nonlinearities in the measurement model.

This paper deals exclusively with finding information
about a finite set of stationary targets. However, ergodic search
generalizes to both time-varying systems as well as estimation of a
continuum of targets (e.g., fields \cite{Cao2013, bender2013}) in a
reasonably straightforward fashion. Field exploration can be achieved by
using an appropriate choice of measurement model and belief update in
the EID calculation \cite{Cao2013, Singh2009, Hoang2014, bender2013,
low2008, souza2014}. Time can be incorporated into the measurement model
describing not just \emph{where} information about a parameter might be
obtained, but also \emph{when}---by extending the state in Section
\ref{trajopt} to use time as a state.

The formulation of ergodic exploration provided in this paper also
assumes that the dynamics are deterministic. However, the determinism
restriction primarily makes calculations and exposition simpler. Adding
stochastic process noise to the model can be achieved by replacing the
deterministic, finite-dimensional equations of motion with the
Fokker-Planck equations \cite{Chirikjian2009} for the nonlinear
stochastic flow, without changing the mathematical formulation of
ergodic control. Moreover, stochastic flows can be efficiently computed
\cite{zhou2003, Wang2002} for a wide variety of robotic problems. Even
when they cannot be, a wide class of stochastic optimal control problems
are easily computable \cite{todorov2005,horowitz2014}, though for
different objectives than ergodicity. Although generalization will be
easier in some cases than others, the generalization of ergodic control
to uncertain stochastic processes may be initially approached rather
procedurally. Generalizing ergodic control to more general uncertain
(non-stochastic) systems, such as robust control strategies
\cite{zhou1998}, would substantially complicate matters and would
require a much more challenging generalization that would be a very
promising avenue of future research.

In addition to the various limiting assumptions mentioned in Sections
\ref{ergoassumptions} and \ref{expassumptions} in constructing the EEDI
algorithm for target localization, one of the major limitations of the
current formulation is computational expense. Computational expense
stems both from the need to calculate a map of the expected information
density over the workspace in order to formulate the ergodic objective
function, and the need to calculate trajectories over a finite time
horizon. The projection-based trajectory optimization involves solving a
set of differential equations, which scale quadratically with the state.
This is not necesserily a problem for applications where offline control
calculations are acceptable, or in a receding horizon framework that
uses efficient numerical methods. To that end, preliminary work has
explored solving a discrete version of ergodic trajectory optimization
using variational integrators \cite{Prabhakar15}. Nevertheless, for
applications that have a linear measurement model, trivial dynamics, and
a simple environment, standard strategies like gradient-based approaches
that only locally approximate the expected information
\cite{Grocholsky06, kreucher2007, lu11, Bourgault02i} would be effective
and much more computationally efficient. The advantage of using ergodic
trajectory optimization is that it is possible to formulate and solve
exploration tasks whether or not the environment is simple or the
measurement model linear, and to perform robustly when these ``nice''
conditions cannot be guaranteed, as in the experimental work featured in
this paper.

\appendix[EEDI for stationary target localization using the
Sensorpod Robot]
 \label{eediappendix}
\subsection{Bayesian Probabilistic Update}\label{prob}
The goal is to estimate a set of $m$ unknown,
 static, parameters $\bm{\theta} =[\theta_1,\theta_2,...,\theta_m]$
 describing individual underwater targets. We assume a measurement $V$
 is made according to a known measurement model
 $V= \Upsilon(\bm \theta, \bm x) + \delta$, where the measurement model
 $\Upsilon(\cdot)$ is a differentiable function of sensor location and target
 parameters, and $\delta$ represents zero mean Gaussian noise. Specifically, we use a previously derived measurement model for submerged, sufficiently isolated, non-conducting spheres \cite{Bai2015}.
The joint distribution
$p(\bm \theta)$ is updated every iteration $k$ of the EEDI algorithm
using a Bayesian filter based on the measurement $V_k(t)$, the
measurement model, and the sensor trajectory $x_k(t)$ over the planning period $T$,
\begin{align} \label{bayes} 
p_{k+1} \left(\bm \theta|   V_k(t) ,\bm x_k(t)\right)= 
  \eta \: p\left(  V_k(t)|\bm \theta,\bm x_k(t)\right)p_{k}(\bm \theta).
\end{align} 
$p(\bm \theta)$ is the PDF calculated at the previous
iteration,
$p_k\left(\bm \theta, V_{k-1}(t) ,\bm x_{k-1}(t)\right)$), $\eta$ is a
normalization factor, and $p(V_k(t)|\bm \theta, \bm x_k(t))$ is the
likelihood function for $\bm \theta$ given $ V_k(t)$.

Assuming independence between individual measurements,
given that the SensorPod state is known and the measurement model is not
time-varying,
the likelihood function for all measurements taken along
$\bm x_k(t)$ is  the product of the likelihood of taking a single
measurement $V_k(t_j)$ at time $t_j$, for all times $t_j\in [t_0,T]$.
Assuming a Gaussian likelihood function, this is
\begin{align}\label{gaussUpdates}
  p( V_k(t)|& \bm \theta, \bm x_k(t) = \\\notag
 &\prod_{j=t_0}^{T}\frac{1}{\sqrt {2\pi}\sigma}
  \exp \left [- \frac{(V_k(t_j)-\Upsilon(\bm \theta,\bm x_k(t_j)))^2}{2
      \sigma^2} \right ].
\end{align}
\subsubsection{Probabilistic Update for Multiple Targets}\label{probmult}
We assume  an additive measurement model $h(\bm { \Theta})$ to describe
the expected measurement for multiple targets,
\begin{align} \label{heq} h(\bm { \Theta},x(t_j)) =& \Upsilon( \bm
  {\theta_1},\bm x(t_j)) + \Upsilon( \bm {\theta_2},\bm x(t_j))+
  ...\\\notag &+ \Upsilon( \bm {\theta_m},\bm x(t_j)),
\end{align}
where $\bm \theta_i$  is an  $m
\times 1$ vector made up of  $m$ parameters  describing
the $i^{th}$ target, and $\bm { \Theta}$ is the $M$ length set of vectors
$[\bm \theta_1,\bm \theta_2,...,\bm \theta_M]$, where $M$ is the
number of targets.
Because we assume the measurements $\Upsilon( \bm {\theta_i})$ for
different targets are independent of each other, we use different
Bayesian filter updates for each target parameter set, evaluating $M$
instances of Eq. \eqref{bayes}. The likelihood function for
parameter set $\bm \theta_i$ is
\begin{align}\notag
  p(V(t)|(\bm \theta_i)) = \prod_{j=1}^{T}
  \frac{1}{\sqrt{2\pi}\sigma}\exp\left [ \frac{(H(\bm \theta_i,\bm
      x(t_j)) - V(t_j))^2}{\sigma^2} \right ]
 \end{align}
where $H(\bm \theta_i,x(t_j))$ is the marginalization \cite{Chirikjian2009} of $h(\bm {
  \Theta},x(t_j))$ over the parameters describing all other targets
$\bm \theta_{j \neq i}$.

\subsection{Expected information density}\label{Fisher Information}

Fisher information \cite{Frie04} defines the informative regions of the
search space based on the measurement model, a function of the parameter
$\theta$. Assuming Gaussian noise, the Fisher information for estimation
of $\theta$ reduces to
\begin{align} 
\label{Fisher2} \mathcal{I}(\theta,x)=\left (\frac{\partial
\Upsilon(\theta,x)}{\partial \theta} \cdot \frac{1}{\sigma} \right ) ^2.
\end{align} 
The Fisher information, $\mathcal I(\theta,x)$ is
the amount of information a measurement provides at location $x$ for a
given estimate of $\theta$ (based on the measurement model).

For estimation of multiple parameters $\bm \theta$ from a random
variable $v$, the Fisher information is represented as an $m \times m$
matrix. For a measurement model $\Upsilon(\bm \theta, \bm x)$ each
element of the Fisher information matrix (FIM) can be simplified to
\begin{align}\label{fishmatrix}
  \mathcal{I}_{i,j}(\bm x, \bm \theta) = \frac{1}{\sigma^2}
  \frac{\partial^2 \Upsilon(\bm \theta, \bm x)}{\partial
    \theta_i \partial\theta_j}.
\end{align}
 \label{fisherguassian} Note that while we assume
Gaussian noise to simplify the expression in \eqref{fishmatrix}, use of
Fisher information in this context does not strictly require Gaussian
noise. The Fisher information can be calculated offline and
stored based on the measurement model which reduces the number of
integrations necessary at each iteration.

Since the estimate of $\bm \theta$ is represented as a
probability distribution function, we  take 
the expected value of each element of $\mathcal{I}(\bm x,
\bm \theta)$ with respect to the joint distribution $p(\bm \theta)$ to
calculate the expected information matrix, $\Phi (\bm x)$. This is
 an $m\times m$ matrix, where the $i,j^{th}$ element is 
\begin{align}\label{Fisher2BIG} 
 \Phi_{i,j}(\bm x)= \frac{1}{\sigma^2}
  \int_{\theta_i} \int_{\theta_j} \frac{\partial^2 \Upsilon( \bm
    \theta,\bm x)}{\partial \theta_i \partial\theta_j} p(\theta_i,
  \theta_j) \, d\theta_j d\theta_i.
\end{align}
This expression can be approximated as a discrete sum as required for
computational efficiency.

 Using the D-optimality metric \cite{emery1998} on the expected
 information matrix, the expected information density (EID) that is
\begin{align}\label{eideq}
EID(\bm x)=\det \Phi(\bm x).
\end{align}

\subsubsection{Expected Information Density  for Multiple Targets}
Since the total information is additive for independent observations, we
can write
\begin{align}\label{inf2}
I(\bm \Theta,x)=I(\bm \theta_1,x)+...+I(\bm \theta_M,x),
\end{align}
where each term is calculated as in \eqref{fishmatrix}.

The expected information density for all parameters, for all targets,
can be calculated as the sum of the determinants of the expected value
of each term in Eq. \eqref{inf2}, $\Phi_i(x)$, given the set of
independent probabilities $p(\bm \theta_i)$ for each set of parameters
$\bm \theta_i$ describing a single target,
\begin{align}\label{eidmult}
EID(x)= \eta \sum_i^M\det \Phi_i(x).
\end{align}
Since the FIM is positive-semidefinite, the determinant of the FIM for
each target is non-negative. Note that this is just one approach of
combining the expected information from several independent sources into
a single map. Another option would have been to calculate the
D-optimality for each target individually and calculate the subsequent
trajectory based only on EID for the target with the highest integrated
information (prior to normalization).


%





\ifCLASSOPTIONcaptionsoff
  \newpage
\fi

\bibliographystyle{IEEEtran}
\bibliography{references}

\begin{thebibliography}{10}
\providecommand{\url}[1]{#1}
\csname url@samestyle\endcsname
\providecommand{\newblock}{\relax}
\providecommand{\bibinfo}[2]{#2}
\providecommand{\BIBentrySTDinterwordspacing}{\spaceskip=0pt\relax}
\providecommand{\BIBentryALTinterwordstretchfactor}{4}
\providecommand{\BIBentryALTinterwordspacing}{\spaceskip=\fontdimen2\font plus
\BIBentryALTinterwordstretchfactor\fontdimen3\font minus
  \fontdimen4\font\relax}
\providecommand{\BIBforeignlanguage}[2]{{%
\expandafter\ifx\csname l@#1\endcsname\relax
\typeout{** WARNING: IEEEtran.bst: No hyphenation pattern has been}%
\typeout{** loaded for the language `#1'. Using the pattern for}%
\typeout{** the default language instead.}%
\else
\language=\csname l@#1\endcsname
\fi
#2}}
\providecommand{\BIBdecl}{\relax}
\BIBdecl

\bibitem{Mathew}
G.~Mathew and I.~Mezic, ``Metrics for ergodicity and design of ergodic dynamics
  for multi-agent system,'' \emph{Physica D-nonlinear Phenomena}, vol. 240, no.
  4-5, pp. 432--442, Feb 2011.

\bibitem{Krah13a}
R.~Krahe and E.~Fortune, ``Electric fishes: neural systems, behavior and
  evolution,'' \emph{J. Exp. Biol.}, vol.~13, 2013.

\bibitem{Nels06a}
M.~E. Nelson and M.~A. MacIver, ``Sensory acquisition in active sensing
  systems,'' \emph{J. Comp. Physiol. {A}}, vol. 192, no.~6, pp. 573--586, 2006.

\bibitem{Neve13a}
I.~D. Neveln, Y.~Bai, J.~B. Snyder, J.~R. Solberg, O.~M. Curet, K.~M. Lynch,
  and M.~A. MacIver, ``{B}iomimetic and bio-inspired robotics in electric fish
  research,'' \emph{J. Exp. Biol.}, vol. 216, no. Pt 13, pp. 2501--2514, Jul
  2013.

\bibitem{Solb08a}
J.~R. Solberg, K.~M. Lynch, and M.~A. MacIver, ``Active electrolocation for
  underwater target localization,'' \emph{International Journal of Robotics
  Research}, vol.~27, no.~5, pp. 529--548, 2008.

\bibitem{MacI04a}
M.~A. MacIver, E.~Fontaine, and J.~W. Burdick, ``Designing future underwater
  vehicles: principles and mechanisms of the weakly electric fish,'' \emph{IEEE
  J. Ocean. Eng.}, vol.~29, no.~3, pp. 651--659, 2004.

\bibitem{Cowen97}
S.~Cowen, S.~Briest, and J.~Dombrowski, ``Underwater docking of autonomous
  undersea vehicles using optical terminal guidance,'' in \emph{{MTS/IEEE}
  {OCEANS} Conf.}, vol.~2, Oct 1997, pp. 1143--1147.

\bibitem{kreucher05s}
C.~Kreucher, K.~Kastella, and A.~O. Hero, ``Sensor management using an active
  sensing approach,'' \emph{Signal Processing}, vol.~85, no.~3, pp. 607--624,
  2005.

\bibitem{Fox98}
D.~Fox, W.~Burgard, and S.~Thrun, ``Active {Markov} localization for mobile
  robots,'' \emph{Robotics and Autonomous Systems}, vol.~25, no. 3-4, pp.
  195--207, 1998.

\bibitem{Cai2009}
C.~Cai and S.~Ferrari, ``Information-driven sensor path planning by approximate
  cell decomposition,'' \emph{IEEE Transactions on Systems, Man, and
  Cybernetics}, vol.~39, no.~3, pp. 672--689, 2009.

\bibitem{Toh06}
J.~Toh and S.~Sukkarieh, ``A {B}ayesian formulation for the prioritized search
  of moving objects,'' in \emph{{IEEE} Int. Conf. on Robotics and Automation
  (ICRA)}, 2006, pp. 219--224.

\bibitem{Cooper08}
J.~Cooper and M.~Goodrich, ``Towards combining {UAV} and sensor operator roles
  in {UAV}-enabled visual search,'' in \emph{{IEEE} Int. Conf. on Human Robot
  Interaction (HRI)}, 2008, pp. 351--358.

\bibitem{hollinger2013}
G.~A. Hollinger, B.~Englot, F.~S. Hover, U.~Mitra, and G.~S. Sukhatme, ``Active
  planning for underwater inspection and the benefit of adaptivity,''
  \emph{International Journal of Robotics Research}, vol.~32, no.~1, pp. 3--18,
  2013.

\bibitem{Denzler02}
J.~Denzler and C.~Brown, ``Information theoretic sensor data selection for
  active object recognition and state estimation,'' \emph{{IEEE} Transactions
  on Pattern Analysis and Machine Intelligence}, vol.~24, no.~2, pp. 145--157,
  2002.

\bibitem{Arbel99}
T.~Arbel and F.~Ferrie, ``Viewpoint selection by navigation through entropy
  maps,'' in \emph{{IEEE} Int. Conf. on Computer Vision}, vol.~1, 1999, pp.
  248--254 vol.1.

\bibitem{Ye99}
Y.~Ye and J.~K. Tsotsos, ``Sensor planning for {3D} object search,''
  \emph{Computer Vision and Image Understanding}, vol.~73, no.~2, pp. 145 --
  168, 1999.

\bibitem{vazquez2001}
P.-P. V{\'a}zquez, M.~Feixas, M.~Sbert, and W.~Heidrich, ``Viewpoint selection
  using viewpoint entropy.'' in \emph{Vision Modeling and Visualization
  Conference}, vol.~1, 2001, pp. 273--280.

\bibitem{massios1998}
N.~A. Massios and R.~B. Fisher, ``A best next view selection algorithm
  incorporating a quality criterion,'' in \emph{British Machine Vision
  Conference}, 1998, pp. 78.1--78.10.

\bibitem{takeuchi1998}
Y.~Takeuchi, N.~Ohnishi, and N.~Sugie, ``Active vision system based on
  information theory,'' \emph{Systems and Computers in Japan}, vol.~29, no.~11,
  pp. 31--39, 1998.

\bibitem{Cao2013}
N.~Cao, K.~H. Low, and J.~M. Dolan, ``Multi-robot informative path planning for
  active sensing of environmental phenomena: A tale of two algorithms,'' in
  \emph{International Conference on Autonomous Agents and Multi-agent Systems},
  2013, pp. 7--14.

\bibitem{bender2013}
A.~Bender, S.~B. Williams, and O.~Pizarro, ``Autonomous exploration of
  large-scale benthic environments,'' in \emph{{IEEE} Int. Conf. on Robotics
  and Automation (ICRA)}, 2013, pp. 390--396.

\bibitem{marchant2014}
R.~Marchant and F.~Ramos, ``{B}ayesian optimisation for informative continuous
  path planning,'' in \emph{{IEEE} Int. Conf. on Robotics and Automation
  (ICRA)}.\hskip 1em plus 0.5em minus 0.4em\relax IEEE, 2014, pp. 6136--6143.

\bibitem{Sim05}
R.~Sim and N.~Roy, ``Global a-optimal robot exploration in {SLAM},'' in
  \emph{{IEEE} Int. Conf. on Robotics and Automation (ICRA)}, 2005, pp.
  661--666.

\bibitem{Leun06a}
C.~Leung, S.~Huang, N.~Kwok, and G.~Dissanayake, ``Planning under uncertainty
  using model predictive control for information gathering,'' \emph{Robotics
  and Autonomous Systems}, vol.~54, no.~11, pp. 898--910, November 2006.

\bibitem{Feder99}
H.~J.~S. Feder, J.~J. Leonard, and C.~M. Smith, ``Adaptive mobile robot
  navigation and mapping,'' \emph{International Journal of Robotics Research},
  vol.~18, no.~7, pp. 650--668, 1999.

\bibitem{VanderHook2012}
J.~Vander~Hook, P.~Tokekar, and V.~Isler, ``Cautious greedy strategy for
  bearing-based active localization: Experiments and theoretical analysis,'' in
  \emph{Robotics and Automation (ICRA), 2012 IEEE International Conference on},
  May 2012, pp. 1787--1792.

\bibitem{Marchant2012}
R.~Marchant and F.~Ramos, ``{B}ayesian optimisation for intelligent
  environmental monitoring,'' in \emph{{IEEE} Int. Conf. on Intelligent Robots
  and Systems (IROS)}, Oct 2012, pp. 2242--2249.

\bibitem{Wong05}
E.-M. Wong, F.~Bourgault, and T.~Furukawa, ``Multi-vehicle {B}ayesian search
  for multiple lost targets,'' in \emph{{IEEE} Int. Conf. on Robotics and
  Automation (ICRA)}, 2005, pp. 3169--3174.

\bibitem{stachniss2003}
C.~Stachniss and W.~Burgard, ``Exploring unknown environments with mobile
  robots using coverage maps,'' in \emph{IJCAI}, 2003, pp. 1127--1134.

\bibitem{kreucher2007}
C.~Kreucher, J.~Wegrzyn, M.~Beauvais, and R.~Conti, ``Multiplatform
  information-based sensor management: an inverted {UAV} demonstration,'' in
  \emph{{SPIE} Defense Transformation and Network-Centric Systems}, vol. 6578,
  2007, pp. 65\,780Y--1--65\,780Y--11.

\bibitem{Roy2006}
N.~Roy and C.~Earnest, ``Dynamic action spaces for information gain
  maximization in search and exploration,'' in \emph{American Controls Conf.
  (ACC)}, June 2006, pp. 6 pp.--.

\bibitem{lu11}
W.~Lu, G.~Zhang, S.~Ferrari, R.~Fierro, and I.~Palunko, ``An information
  potential approach for tracking and surveilling multiple moving targets using
  mobile sensor agents,'' in \emph{{SPIE} Unmanned Systems Technology}, vol.
  8045, 2011, pp. 80\,450T--1--80\,450T--13.

\bibitem{Bourgault02i}
F.~Bourgault, A.~A. Makarenko, S.~Williams, B.~Grocholsky, and
  H.~Durrant-Whyte, ``Information based adaptive robotic exploration,'' in
  \emph{{IEEE} Int. Conf. on Intelligent Robots and Systems (IROS)}, vol.~1,
  2002, pp. 540 -- 545.

\bibitem{Elfes89}
A.~Elfes, ``Using occupancy grids for mobile robot perception and navigation,''
  \emph{Computer}, vol.~22, no.~6, pp. 46 --57, Jun 1989.

\bibitem{Singh2009}
A.~Singh, A.~Krause, C.~Guestrin, and W.~J. Kaiser, ``Efficient informative
  sensing using multiple robots,'' \emph{In Journal of Journal of Artificial
  Intelligence Research (JAIR)}, vol.~34, pp. 707--755, 2009.

\bibitem{Hoang2014}
T.~N. Hoang, K.~H. Low, P.~Jaillet, and M.~Kankanhalli, ``Nonmyopic
  $\epsilon$-{Bayes}-optimal active learning of gaussian processes,'' in
  \emph{International Conference on Machine Learning}, 2014, pp. 739--747.

\bibitem{low2008}
K.~H. Low, J.~M. Dolan, and P.~Khosla, ``Adaptive multi-robot wide-area
  exploration and mapping,'' in \emph{Conference on Autonomous agents and
  multiagent systems}.\hskip 1em plus 0.5em minus 0.4em\relax International
  Foundation for Autonomous Agents and Multiagent Systems, 2008, pp. 23--30.

\bibitem{souza2014}
J.~Souza, R.~Marchant, L.~Ott, D.~Wolf, and F.~Ramos, ``{Bayesian} optimisation
  for active perception and smooth navigation,'' in \emph{{IEEE} Int. Conf. on
  Robotics and Automation (ICRA)}, May 2014, pp. 4081--4087.

\bibitem{Bajcsy88}
R.~Bajcsy, ``Active perception,'' \emph{Proceedings of the IEEE}, vol.~76,
  no.~8, pp. 996--1005, 1988.

\bibitem{spletzer03}
J.~R. Spletzer and C.~J. Taylor, ``Dynamic sensor planning and control for
  optimally tracking targets,'' \emph{International Journal of Robotics
  Research}, vol.~22, no.~1, pp. 7--20, 2003.

\bibitem{dasgupta2006}
B.~DasGupta, J.~P. Hespanha, J.~Riehl, and E.~Sontag, ``Honey-pot constrained
  searching with local sensory information,'' \emph{Nonlinear Analysis: Theory,
  Methods \& Applications}, vol.~65, no.~9, pp. 1773--1793, 2006.

\bibitem{zhang09}
G.~Zhang and S.~Ferrari, ``An adaptive artificial potential function approach
  for geometric sensing,'' in \emph{{IEEE} Int. Conf. on Decision and Control
  (CDC)}, 2009, pp. 7903--7910.

\bibitem{hager91}
G.~Hager and M.~Mintz, ``Computational methods for task-directed sensor data
  fusion and sensor planning,'' \emph{International Journal of Robotics
  Research}, vol.~10, no.~4, pp. 285--313, 1991.

\bibitem{Benet02}
G.~Benet, F.~Blanes, J.~Sim{\'o}, and P.~P{\'e}rez, ``Using infrared sensors
  for distance measurement in mobile robots,'' \emph{Robotics and Autonomous
  Systems}, vol.~40, no.~4, pp. 255 -- 266, 2002.

\bibitem{Denzler03}
J.~Denzler, M.~Zobel, and H.~Niemann, ``Information theoretic focal length
  selection for real-time active 3d object tracking,'' in \emph{{IEEE} Int.
  Conf. on Computer Vision}, vol.~1, Oct 2003, pp. 400--407.

\bibitem{Tisd09a}
J.~Tisdale, Z.~Kim, and J.~K. Hedrick, ``Autonomous {UAV} path planning and
  estimation,'' \emph{{IEEE} Robotics and Automation Magazine}, vol.~16, no.~2,
  pp. 35--42, June 2009.

\bibitem{Grocholsky06}
B.~Grocholsky, J.~Keller, V.~Kumar, and G.~Pappas, ``Cooperative air and ground
  surveillance,'' \emph{{IEEE} Robotics and Automation Magazine}, vol.~13,
  no.~3, pp. 16--25, 2006.

\bibitem{lu2014}
W.~Lu, G.~Zhang, and S.~Ferrari, ``An information potential approach to
  integrated sensor path planning and control,'' \emph{{IEEE} Transactions on
  Robotics}, vol.~30, no.~4, pp. 919--934, Aug 2014.

\bibitem{zhang09B}
G.~Zhang, S.~Ferrari, and M.~Qian, ``An information roadmap method for robotic
  sensor path planning,'' \emph{Journal of Intelligent and Robotic Systems},
  vol.~56, no. 1-2, pp. 69--98, 2009.

\bibitem{liao04}
X.~Liao and L.~Carin, ``Application of the theory of optimal experiments to
  adaptive electromagnetic-induction sensing of buried targets,'' \emph{{IEEE}
  Transactions on Pattern Analysis and Machine Intelligence}, vol.~26, no.~8,
  pp. 961--972, 2004.

\bibitem{emery1998}
A.~Emery and A.~V. Nenarokomov, ``Optimal experiment design,''
  \emph{Measurement Science and Technology}, vol.~9, no.~6, p. 864, 1998.

\bibitem{Ucinski1999}
D.~Ucinski and J.~Korbicz, ``Path planning for moving sensors in parameter
  estimation of distributed systems,'' in \emph{Workshop on Robot Motion and
  Control (RoMoCo)}, 1999, pp. 273--278.

\bibitem{Ucinski2000}
D.~Ucinski, ``Optimal sensor location for parameter estimation of distributed
  processes,'' \emph{International Journal of Control}, vol.~73, no.~13, pp.
  1235--1248, 2000.

\bibitem{Frie04}
R.~B. Frieden, \emph{Science from Fisher Information: A Unification}.\hskip 1em
  plus 0.5em minus 0.4em\relax Cambridge University Press, 2004.

\bibitem{Atanasov2014}
N.~Atanasov, B.~Sankaran, J.~Le~Ny, G.~Pappas, and K.~Daniilidis, ``Nonmyopic
  view planning for active object classification and pose estimation,''
  \emph{{IEEE} Transactions on Robotics}, vol.~30, no.~5, pp. 1078--1090, Oct
  2014.

\bibitem{Li05}
Y.~F. Li and Z.~G. Liu, ``Information entropy based viewpoint planning for
  {3-D} object reconstruction,'' \emph{{IEEE} Transactions on Robotics},
  vol.~21, no.~3, pp. 324--327, 2005.

\bibitem{Rahimi2005}
M.~Rahimi, M.~Hansen, W.~Kaiser, G.~Sukhatme, and D.~Estrin, ``Adaptive
  sampling for environmental field estimation using robotic sensors,'' in
  \emph{{IEEE} Int. Conf. on Intelligent Robots and Systems (IROS)}, Aug 2005,
  pp. 3692--3698.

\bibitem{Hollinger2014}
G.~A. Hollinger and G.~S. Sukhatme, ``Sampling-based robotic information
  gathering algorithms,'' \emph{International Journal of Robotics Research},
  vol.~33, no.~9, pp. 1271--1287, 2014.

\bibitem{Ryan2010}
A.~Ryan and J.~K. Hedrick, ``Particle filter based information-theoretic active
  sensing,'' \emph{Robotics and Autonomous Systems}, vol.~58, no.~5, pp. 574 --
  584, 2010.

\bibitem{MartinezCantin2009}
R.~Martinez-Cantin, N.~de~Freitas, E.~Brochu, J.~Castellanos, and A.~Doucet,
  ``\BIBforeignlanguage{English}{A {B}ayesian exploration-exploitation approach
  for optimal online sensing and planning with a visually guided mobile
  robot},'' \emph{\BIBforeignlanguage{English}{Autonomous Robots}}, vol.~27,
  no.~2, pp. 93--103, 2009.

\bibitem{Jacobs}
H.~Jacobs, S.~Nair, and J.~Marsden, ``Multiscale surveillance of {Riemannian}
  manifolds,'' in \emph{American Controls Conf. (ACC)}, 2010, pp. 5732--5737.

\bibitem{wilson2014}
A.~D. Wilson, J.~A. Schultz, and T.~D. Murphey, ``Trajectory synthesis for
  fisher information maximization,'' \emph{{IEEE} Transactions on Robotics},
  vol.~30, no.~6, 2014.

\bibitem{Song12}
D.~Song, C.-Y. Kim, and J.~Yi, ``Simultaneous localization of multiple unknown
  and transient radio sources using a mobile robot,'' \emph{{IEEE} Transactions
  on Robotics}, vol.~28, no.~3, pp. 668 --680, june 2012.

\bibitem{Kim2014}
C.-Y. Kim, D.~Song, Y.~Xu, J.~Yi, and X.~Wu, ``Cooperative search of multiple
  unknown transient radio sources using multiple paired mobile robots,''
  \emph{{IEEE} Transactions on Robotics}, vol.~30, no.~5, pp. 1161--1173, Oct
  2014.

\bibitem{miller13R}
{L.M. Miller and T.D. Murphey}, ``Trajectory optimization for continuous
  ergodic exploration,'' in \emph{American Controls Conf. (ACC)}, 2013, pp.
  4196--4201.

\bibitem{miller13SE}
L.~M. Miller and T.~D. Murphey, ``Trajectory optimization for continuous
  ergodic exploration on the motion group {SE(2)},'' in \emph{{IEEE} Int. Conf.
  on Decision and Control (CDC)}, 2013, p. In Press.

\bibitem{Acar2003}
E.~Acar, H.~Choset, Y.~Zhang, and M.~Schervish, ``Path planning for robotic
  demining: Robust sensor-based coverage of unstructured environments and
  probabilistic methods,'' \emph{International Journal of Robotics Research},
  vol.~22, no. 7 - 8, pp. 441 -- 466, July 2003.

\bibitem{choset2001}
H.~Choset, ``Coverage for robotics--a survey of recent results,'' \emph{Annals
  of mathematics and artificial intelligence}, vol.~31, no. 1-4, pp. 113--126,
  2001.

\bibitem{Krause2007}
A.~Krause and C.~Guestrin, ``Nonmyopic active learning of gaussian processes:
  an exploration-exploitation approach,'' in \emph{International Conference on
  Machine learning}.\hskip 1em plus 0.5em minus 0.4em\relax ACM, 2007, pp.
  449--456.

\bibitem{Hauser}
J.~Hauser, ``A projection operator approach to the optimization of trajectory
  functionals,'' in \emph{IFAC world congress}, vol.~4, 2002, pp. 3428--3434.

\bibitem{Chirikjian2009}
G.~S. Chirikjian, \emph{Stochastic Models, Information Theory, and Lie Groups,
  Volume 1: Classical Results and Geometric Methods}.\hskip 1em plus 0.5em
  minus 0.4em\relax Springer Science \& Business Media, 2009.

\bibitem{Chirikjian}
G.~S. Chirikjian and A.~B. Kyatkin, \emph{Engineering Applications of
  Noncommutative Harmonic Analysis}.\hskip 1em plus 0.5em minus 0.4em\relax CRC
  Press, 2000.

\bibitem{Bai2015}
Y.~Bai, J.~B. Snyder, M.~Peshkin, and M.~A. MacIver, ``Finding and identifying
  simple objects underwater with active electrosense,'' \emph{International
  Journal of Robotics Research}, vol.~1, no.~23, 2015.

\bibitem{Thrun2003}
S.~Thrun, ``Learning occupancy grid maps with forward sensor models,''
  \emph{Autonomous Robots}, vol.~15, no.~2, pp. 111--127, 2003.

\bibitem{thrun05p}
S.~Thrun, W.~Burgard, and D.~Fox, \emph{Probabilistic Robotics}.\hskip 1em plus
  0.5em minus 0.4em\relax Cambridge, Mass: The MIT Press, 2005.

\bibitem{silverman13}
Y.~Silverman, L.~M. Miller, M.~A. MacIver, and T.~D. Murphey, ``Optimal
  planning for information acquisition,'' in \emph{{IEEE} Int. Conf. on
  Intelligent Robots and Systems (IROS)}, vol. 5974- 5980, 2013.

\bibitem{Nelson2006}
M.~E. Nelson and M.~A. MacIver, ``Sensory acquisition in active sensing
  systems,'' \emph{Journal of Comparative Physiology A}, vol. 192, no.~6, pp.
  573--586, 2006.

\bibitem{zhou2003}
Y.~Zhou and G.~S. Chirikjian, ``Probabilistic models of dead-reckoning error in
  nonholonomic mobile robots,'' in \emph{{IEEE} Int. Conf. on Robotics and
  Automation (ICRA)}, vol.~2, 2003, pp. 1594--1599.

\bibitem{Wang2002}
Y.~Wang and G.~Chirikjian, ``A diffusion-based algorithm for workspace
  generation of highly articulated manipulators,'' in \emph{{IEEE} Int. Conf.
  on Robotics and Automation (ICRA)}, vol.~2, 2002, pp. 1525--1530 vol.2.

\bibitem{todorov2005}
E.~Todorov and W.~Li, ``A generalized iterative lqg method for locally-optimal
  feedback control of constrained nonlinear stochastic systems,'' in
  \emph{American Controls Conf. (ACC)}, 2005, pp. 300--306.

\bibitem{horowitz2014}
M.~B. Horowitz and J.~W. Burdick, ``Semidefinite relaxations for stochastic
  optimal control policies,'' in \emph{American Controls Conf. (ACC)}, 2014,
  pp. 3006--3012.

\bibitem{zhou1998}
K.~Zhou and J.~C. Doyle, \emph{Essentials of robust control}.\hskip 1em plus
  0.5em minus 0.4em\relax Prentice hall Upper Saddle River, NJ, 1998, vol. 180.

\bibitem{Prabhakar15}
A.~Prabhakar, K.~Fla{\ss}kamp, and T.~Murphey, ``Time discretization in optimal
  ergodic control,'' in \emph{{IEEE} Int. Conf. on Decision and Control (CDC)},
  2015.

\end{thebibliography}



%



%

\end{document}